%% file: main.tex
\documentclass[pmlr]{jmlr}

\usepackage{mathtools}

\usepackage{booktabs}
\usepackage{multirow}

\usepackage{microtype}
\usepackage{xcolor}
\usepackage{enumitem}

\jmlrprehyperref{\PassOptionsToPackage{colorlinks=true,linkcolor=blue!70!black,citecolor=blue!70!black,urlcolor=blue!70!black}{hyperref}}

\newcommand{\vimp}{V(\text{imp})}
\newcommand{\vrl}{V(\pi_{\text{RL}})}
\newcommand{\vphy}{V(\pi_{\text{physician}})}
\newcommand{\cmark}{\checkmark}
\newcommand{\xmark}{$\times$}

\newcommand{\rstd}{\ensuremath{r_{\text{std}}}}
\newcommand{\rend}{\ensuremath{r_{\text{END}}}}
\newcommand{\rdml}{\ensuremath{r_{\text{DML}}}}
\newcommand{\rfull}{\ensuremath{r_{\text{full}}}}
\newcommand{\ryr}{\ensuremath{r_{\text{1yr}}}}
\newcommand{\ryrdml}{\ensuremath{r_{\text{1yr,DML}}}}
\newcommand{\rendprime}{\ensuremath{r'_{\text{END}}}}

\jmlrproceedings{PMLR}{Proceedings of Machine Learning Research}
\jmlrvolume{340}
\jmlryear{2026}
\jmlrworkshop{Machine Learning for Healthcare}

\begin{document}

\title[Reward-Embedded Confounding in Clinical Offline RL]{Confounding Masquerading as Improvement: A Systematic Evaluation of Offline Reinforcement
       Learning for Stroke Antithrombotic Treatment in a 129,000-Patient Registry}

\author{\Name{Kihun Rhee}
        \Email{rheekh00@snu.ac.kr}\\
        \addr AI Research Center, JLK Inc., Seoul, South Korea}

\maketitle

\begin{abstract}
Recent offline reinforcement learning (RL) studies report data-driven policies
outperform physician decisions by 10--32\,\% on clinical outcomes.
We conduct a systematic, partially crossed evaluation of five offline RL
algorithm families and 14 reward designs
on 44,894 post-2018 acute ischemic stroke patients
from the nationwide stroke registry ($N = 129{,}033$).

Standard Fitted Q-Evaluation yields
$\vimp = +0.0069$ ($p = 0.048$); adding an Early Neurological Deterioration
penalty strengthens the signal to $+0.0101$ ($p = 0.0002$,
approximate E-value $= 18.37$) --- estimates that could support a premature
positive policy-improvement claim despite relying on a confounded reward
function.

We identify \emph{reward-embedded confounding}, where the proxy terminal reward
encodes baseline severity/prognosis information in addition to any treatment
efficacy signal. E-value analysis does not evaluate this pathway because the
confounding is carried through the reward channel. A $2{\times}2$ factorial
experiment reveals that terminal reward confounding alone more than eliminates
the apparent improvement,
accounting for 218.6\,\% of the $\rend \to \rfull$ signal change
(i.e., removing terminal confounding overshoots null).

After DML-inspired GBM reward residualization, $\vimp$ attenuates to
$+0.0033$ ($p = 0.132$); full deconfounding yields $+0.0025$ ($p = 0.291$).
Three independent approaches converge away from a clinically meaningful
aggregate policy-improvement claim: FQE-based reward deconfounding, a T-learner
showing 75--90\,\% attenuation with a clinically small residual
($<6\,\%$ MCID), and direct ischemic-stroke recurrence analysis (IPW/AIPW).
A within-registry 1-year mRS factorial ($N{=}35{,}744$) replicates the pattern:
$\vimp = +0.0133$ attenuates to $-0.0004$ after reward residualization
(103.3\,\% attenuation), indicating that the finding is not specific to the
3-month mRS horizon in this registry. We present an empirically motivated
6-step evaluation checklist; applied retrospectively, three highly cited
clinical RL papers do not report the FQE confirmation specified by Step~1.

Despite the overall null, National Institutes of Health Stroke Scale
(NIHSS)-stratified heterogeneity
(T-learner conditional average treatment effect (CATE) ratio $4.6\times$,
corroborated by Q-evaluation $3.4\times$ gradient)
identifies a hypothesis-generating subgroup for prospective trial design;
hospital-level disagreement did not persist after full reward deconfounding.
\end{abstract}



\input{introduction}

\input{methods}

\input{results}

\input{protocol}

\input{discussion}


\section*{Data and Code Availability}
Individual-level CRCS-K data cannot be publicly deposited because of informed consent,
ethics, data-sharing, privacy, and regulatory restrictions. Qualified researchers
may request a de-identified minimum dataset for a legitimate academic proposal,
subject to CRCS-K Steering Committee and institutional approvals; the author cannot
independently redistribute it. An artifact package is being prepared for a planned
arXiv/PMLR release, including code, configurations, figure scripts, and synthetic
examples for implementation inspection. The package will not include individual-level
CRCS-K data and cannot reconstruct the cohort or reproduce patient-level results
without approved data access.


\bibliography{references}


\acks{%
No external funding was received. This work was conducted as part of the
author's employment at JLK Inc.
The author is an employee of JLK Inc., a medical AI company.
Data access was provided under the CRCS-K multicenter research agreement.
}


\appendix

\clearpage
\section{Cohort Demographics}
\label{app:cohort}
\input{tab_cohort}

\section{Temporal Cohort Restriction}
\label{app:temporal}
The primary RL cohort ($N = 44{,}894$) is derived from the full CRCS-K
registry ($N = 129{,}033$) through the following steps:

\begin{enumerate}[nosep, leftmargin=*]
  \item \textbf{Ischemic stroke episodes}: $129{,}033$ consecutive
    admissions across 20 hospitals (2008--2025) $\to$ $116{,}215$
    complete 3-step MDP episodes after removing records with missing
    acute or discharge treatment data.
  \item \textbf{Post-2018 temporal restriction}: $116{,}215$ $\to$
    ${\approx}49{,}800$ episodes with admission date $\geq$ 2018-04-05
    (following the 2018 Korean Stroke Society guideline update that
    standardized dual antiplatelet therapy (DAPT) for non-cardioembolic
    stroke and expanded non-vitamin K oral anticoagulant (NOAC)
    indications for AF).
  \item \textbf{Valid 3-month mRS}: ${\approx}49{,}800$ $\to$ $44{,}895$
    episodes with non-missing 3-month modified Rankin Scale
    (${\approx}4{,}900$ excluded for missing mRS, who have higher mean
    NIHSS $= 6.8$ vs.\ $4.0$, suggesting mild missing-not-at-random (MNAR) bias).
  \item \textbf{Complete 2-step MDP episodes}: $44{,}895$ $\to$
    $44{,}894$ after removing 1 patient lacking complete episode data
    for the 2-step (acute $\to$ discharge) formulation.
\end{enumerate}

\noindent
Including pre-2018 data introduces guideline-era confounding: treatment patterns
differ across eras for reasons unrelated to treatment efficacy, biasing
the reward signal.
Temporal stability analysis shows that a CQL model
trained on pre-2018 data achieves only 65.4\,\% action agreement on
post-2018 holdout patients, confirming distribution shift.

\section{Positivity Analysis Details}
\label{app:positivity}
The effective sample size (ESS) for importance-sampling OPE is defined as
\[
  \text{ESS} = \Bigl(\textstyle\sum_i w_i\Bigr)^2 \Big/ \textstyle\sum_i w_i^2
\]
(Kish's effective sample size~\citep{kish1965}),
where $w_i$ are the cumulative importance weights.
In Phase 8B (\rstd, CQL $\alpha{=}4.0$), ESS $= 93.6$ patients
(out of $44{,}894$, ESS ratio $= 0.21\,\%$).
Cumulative IS weight maximum reached $9.8 \times 10^{14}$.
A commonly recommended practical threshold for reliable IS-based OPE is
ESS $\geq 10\,\%$~\citep{kang2007}; our ESS is $48\times$ below this threshold.
Under the undeconfounded \rstd{} reward, clip-to-10 weighted doubly robust
(WDR) estimation reduces weight variance
but produces $|\vimp_{\text{DR}} - \vimp_{\text{FQE}}| = 0.055$, far
exceeding the 0.02 agreement threshold, so this diagnostic is not met for
\rstd{}.
Under the deconfounded \rdml{} reward, the same statistic is 0.007, satisfying
the Step~2 agreement diagnostic (\S\ref{sec:results:act2}).

\clearpage
\section{Experiment Overview: Evidence Map}
\label{app:evidence_map}

\begin{figure}[!ht]
  \centering
  \makebox[\textwidth][c]{%
    \includegraphics[width=1.35\textwidth, keepaspectratio]{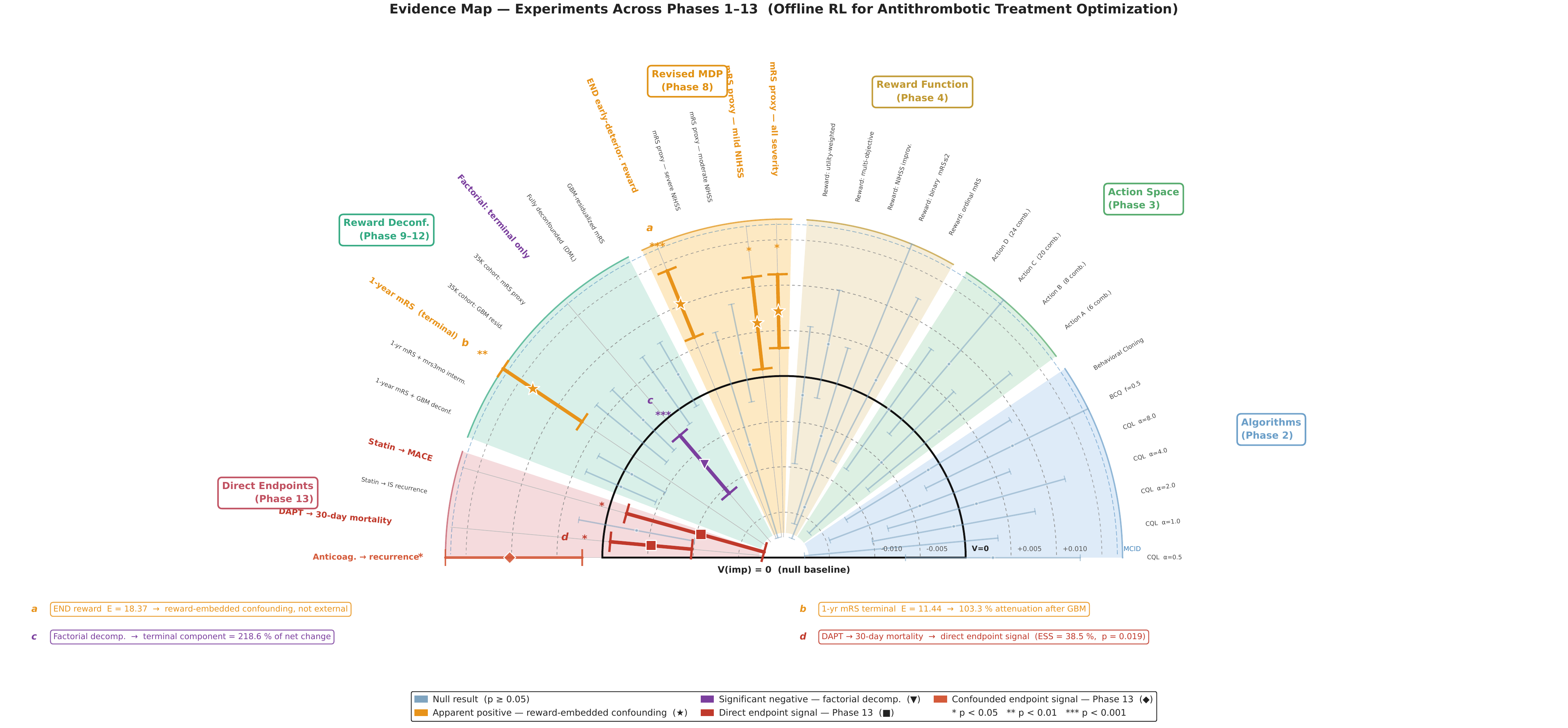}}
  \caption{%
    \textbf{Evidence map across Phases 1--13.}
    Each arc represents one experimental comparison; radial position encodes
    $\vimp$ (positive outward, negative inward), and color indicates result
    category. Dashed arc: $\vimp = 0$ (physician baseline).
    Table~\ref{tab:evidence_phase_legend} maps each phase label to its cohort,
    reward or analysis family, and diagnostic purpose. The map shows that
    positive signals are concentrated in non-deconfounded reward designs and
    attenuate after GBM residualization, visually summarizing the
    reward-embedded confounding diagnostic sequence.
  }
  \label{fig:evidence_map}
\end{figure}

Figure~\ref{fig:evidence_map} provides a radial overview of experiments across
Phases~1--13. Each arc encodes $\vimp$ (positive outward, negative inward), and
color indicates whether the result is null, an apparent positive signal under a
non-residualized reward, or a direct causal-analysis result. Positive signals
cluster in non-deconfounded reward designs and attenuate after GBM
residualization, visually summarizing the reward-embedded confounding
diagnostic sequence.

\begin{table}[!ht]
  \caption{Phase legend for Figure~\ref{fig:evidence_map}. The table maps the
  phase labels used in the evidence map to the cohort, reward or analysis family,
  and diagnostic purpose.}
  \label{tab:evidence_phase_legend}
  \centering
  \footnotesize
  \setlength{\tabcolsep}{3pt}
  \begin{tabular}{@{}p{1.7cm}p{3.0cm}p{4.0cm}p{5.7cm}@{}}
    \toprule
    Phase(s) & Cohort & Reward / analysis & Purpose \\
    \midrule
    1--2 & Initial RL cohorts & Algorithm and OPE screening &
      Select candidate algorithms and quantify direct-Q versus FQE overestimation. \\
    3 & Initial RL cohorts & Action-space variants &
      Test whether finer antithrombotic action encodings change policy-value conclusions. \\
    4 & Initial RL cohorts & Terminal reward variants &
      Assess reward-formulation sensitivity before reward deconfounding. \\
    5--7 & Diagnostic subsets & Subgroup, temporal, E-value, and support diagnostics &
      Identify heterogeneity, guideline-era instability, external-confounding sensitivity, and IS support limits. \\
    8 & Post-2018, $N{=}44{,}894$ & \rstd{} standard FQE and NIHSS strata &
      Establish the revised 2-step MDP baseline and the raw-positive policy-improvement estimate. \\
    9 & Post-2018, $N{=}44{,}894$ & \rdml{} prognosis-residualized terminal reward &
      Stress-test whether the standard positive signal persists after terminal reward residualization. \\
    10 & Post-2018, $N{=}44{,}894$ & \rend{}, \rendprime{}, \rfull{} &
      Decompose terminal and END-channel confounding with the $2{\times}2$ factorial and waterfall. \\
    11--12 & 1-year follow-up, $N{=}35{,}744$ & \ryr{} and \ryrdml{} &
      Test whether the raw-positive, residualized-null pattern appears beyond 3-month mRS. \\
    13 & Direct endpoint cohorts & IPW/AIPW recurrence and safety endpoints &
      Triangulate the RL reward findings against treatment-aligned clinical endpoints. \\
    \bottomrule
  \end{tabular}
\end{table}

\section{Algorithm $\times$ Reward Interaction}
\label{app:heatmap}

The experiment grid was partially rather than fully crossed. We first screened
five algorithm families under a common initial-cohort reward, then performed the
post-2018 reward-deconfounding factorial with the prespecified primary CQL
configuration. Tables~\ref{tab:algorithm_screen_full} and
\ref{tab:reward_screen_full} report the evaluated cells explicitly; no values
are imputed for untested algorithm--reward combinations.

\begin{table}[!ht]
  \caption{Full common-reward algorithm screen. FQE is reported for CQL, BCQ,
  IQL, and BC; DT has a direct value estimate because FQE was unavailable.
  Each configuration has 15 seed--fold evaluations, and none shows positive
  improvement evidence.}
  \label{tab:algorithm_screen_full}
  \centering
  \footnotesize
  \setlength{\tabcolsep}{4pt}
  \begin{tabular}{@{}llccp{5.0cm}@{}}
    \toprule
    Family & Configuration & $\vimp$ & 95\,\% CI & Interpretation \\
    \midrule
    CQL & $\alpha=0.5$ & $+0.0025$ & $[-0.0071,+0.0122]$ & Interval includes zero \\
    CQL & $\alpha=1.0$ & $-0.0088$ & $[-0.0215,+0.0040]$ & Interval includes zero \\
    CQL & $\alpha=2.0$ & $-0.0012$ & $[-0.0104,+0.0080]$ & Interval includes zero \\
    CQL & $\alpha=4.0$ & $+0.0021$ & $[-0.0081,+0.0123]$ & Interval includes zero \\
    CQL & $\alpha=8.0$ & $-0.0038$ & $[-0.0146,+0.0070]$ & Interval includes zero \\
    BCQ & $f=0.5$ & $+0.0081$ & $[-0.0027,+0.0188]$ & Interval includes zero \\
    IQL & $\tau=0.7$ & $-0.0117$ & $[-0.0222,-0.0012]$ & Below physician value \\
    DT & Default & $0.0000$ & $[0.0000,0.0000]$ & Direct-value collapse; no FQE \\
    BC & Default & $-0.0014$ & $[-0.0120,+0.0091]$ & Interval includes zero \\
    \bottomrule
  \end{tabular}
\end{table}

\begin{table}[!ht]
  \caption{Full verified reward-stress-test cells for CQL $\alpha=4.0$.
  Preliminary rewards use the initial cohort/default MDP; named rewards use the
  post-2018 two-step cohort. The raw-positive estimates do not persist after
  prognosis residualization.}
  \label{tab:reward_screen_full}
  \centering
  \footnotesize
  \setlength{\tabcolsep}{3pt}
  \resizebox{\textwidth}{!}{%
  \begin{tabular}{@{}p{2.3cm}p{5.2cm}ccp{3.8cm}@{}}
    \toprule
    Cohort & Reward design & $\vimp$ & 95\,\% CI & Interpretation \\
    \midrule
    Initial & Ordinal mRS terminal & $+0.0021$ & $[-0.0081,+0.0123]$ & Interval includes zero \\
    Initial & Binary mRS $\leq2$ & $-0.0140$ & $[-0.0453,+0.0173]$ & Interval includes zero \\
    Initial & NIHSS-change terminal & $-0.0063$ & $[-0.0166,+0.0039]$ & Interval includes zero \\
    Initial & Multi-objective composite & $+0.0042$ & $[-0.0020,+0.0104]$ & Interval includes zero \\
    Initial & mRS utility weights & $-0.0020$ & $[-0.0095,+0.0055]$ & Interval includes zero \\
    \midrule
    Post-2018 & \rstd{} standard reward & $+0.0069$ & $[+0.0001,+0.0140]$ & Borderline raw-positive \\
    Post-2018 & \rend{} with raw END & $+0.0101$ & $[+0.0058,+0.0145]$ & Raw-positive, confounded \\
    Post-2018 & \rdml{} residual terminal & $+0.0033$ & $[-0.0011,+0.0077]$ & Interval includes zero \\
    Post-2018 & \rendprime{} residual terminal/raw END & $-0.0065$ & $[-0.0111,-0.0019]$ & Factorial diagnostic cell \\
    Post-2018 & \rfull{} residual terminal and END & $+0.0025$ & $[-0.0024,+0.0075]$ & Interval includes zero \\
    1-year & \ryr{} raw terminal & $+0.0133$ & $[+0.0062,+0.0205]$ & Raw-positive \\
    1-year & \ryrdml{} residual terminal & $-0.0004$ & $[-0.0050,+0.0040]$ & Interval includes zero \\
    \bottomrule
  \end{tabular}
  }
\end{table}

\section{Instrumental Variable Details}
\label{app:iv}
We used hospital-level anticoagulation (AC) prescription rate as instrument
for AC treatment in the AF subgroup ($N = 6{,}458$).
The instrument is strong (first-stage $F = 515.6$) but \textbf{fails the exclusion
restriction}: 10 of 28 covariate balance checks show significant imbalance
($p < 0.05$), indicating that hospital AC rate correlates with other
institutional practices that directly affect mRS outcomes.
Because the exclusion restriction is violated, the IV estimate is not
causally interpretable; we report it for completeness:
LATE $= +0.279$ ($p = 0.712$, 95\,\%~CI $[-1.06, +1.21]$),
directionally consistent with the null but unreliable.
Endogeneity of treatment is confirmed (2SRI residual test $p < 0.001$),
validating the need for causal methods beyond naive OPE.

\section{Difference-in-Differences Details}
\label{app:did}
We use the following abbreviations: two-way fixed effects (TWFE), average treatment effect on the treated (ATT), wild cluster bootstrap (WCB), and randomization inference (RI).

\textbf{DAPT experiment}: 39,947 patients, 10 sites, staggered DAPT adoption
2010--2020. TWFE ATT $= +0.134$ ($p = 0.170$); wild cluster bootstrap (WCB)
$p = 0.247$; randomization inference $p = 0.438$.
Parallel trends violated ($F = 20.58$, $p < 0.001$) --- results are exploratory.

\textbf{NOAC experiment}: 13,387 AF patients, 15 sites, NOAC adoption
2015--2016. TWFE ATT $= +0.211$ ($p = 0.071$); WCB $p = 0.104$.
Binary mRS$\leq$2 outcome: TWFE $p = 0.021$, WCB $p = 0.110$,
RI $p = 0.757$ --- estimate unstable across inference methods.
Seven sensitivity checks: 3 pass (placebo, binary, statin),
2 mixed (age stratification), 2 fail (confounding robustness).
E-value for DAPT $= 1.36$; for NOAC $= 1.43$ --- modest confounding
suffices to explain both estimates.

\section{Direct Recurrence Analysis}
\label{app:recurrence}

We performed IPW/AIPW treatment effect estimation and cause-specific Cox
regression on stroke recurrence as a direct clinical endpoint ($N = 32{,}583$
post-2018 ischemic stroke patients with 3-month follow-up).
Table~\ref{tab:recurrence} summarizes results across three binary comparisons
and four outcome endpoints.

\begin{table}[h]
\centering
\footnotesize
\caption{Direct recurrence analysis: AIPW ATE and IPW-weighted cause-specific Cox HR
  (Cox HR computed only for death endpoint where PH assumption holds).
  Bold: $p < 0.05$. IS = ischemic stroke.
  MACE = major adverse cardiovascular events. --- = Cox not computed.}
\label{tab:recurrence}
\begin{tabular}{@{}llrrrr@{}}
  \toprule
  Comparison & Outcome & ATE & $p$ & HR & $p$ \\
  \midrule
  C1: Statin & IS recur.   & $-0.003$ & .233 & --- & --- \\
             & Any recur.  & $-0.004$ & .284 & --- & --- \\
             & Death       & $-0.005$ & .078 & 0.870 & $<$.001 \\
             & MACE        & $-0.010$ & .021 & --- & --- \\
  \midrule
  C2: DAPT   & IS recur.   & $+0.003$ & .168 & --- & --- \\
             & Any recur.  & $+0.004$ & .174 & --- & --- \\
             & Death       & $\mathbf{-0.005}$ & \textbf{.019} & \textbf{0.938} & \textbf{.001} \\
             & MACE        & $-0.001$ & .824 & --- & --- \\
  \midrule
  C3: AC (AF) & IS recur.  & $+0.007$ & .117 & --- & --- \\
              & Any recur. & $\mathbf{+0.010}$ & \textbf{.019} & --- & --- \\
              & Death      & $-0.001$ & .861 & 0.922 & .208 \\
              & MACE       & $+0.006$ & .495 & --- & --- \\
  \bottomrule
\end{tabular}
\end{table}

Key findings:
(1)~No comparison achieves significant ischemic stroke recurrence reduction.
(2)~DAPT shows the most robust signal: mortality reduction (HR~$= 0.938$,
$p = 0.001$), consistent with CHANCE/POINT trial evidence~\citep{wang2013chance,johnston2018point}.
(3)~\textbf{Caution:} The AC contrast in the AF subgroup showed increased
any-cause recurrence (ATE $= +0.010$, $p = 0.019$); this should \emph{not}
be interpreted as evidence that anticoagulation increases recurrence risk.
The likely explanation is confounding by indication: AF patients receiving AC
have higher stroke severity (mean NIHSS 6.8 vs.\ 4.2, SMD $> 0.1$)
and early hemorrhagic risk~\citep{paciaroni2015}, and the small control group
($n = 527$) provides insufficient support for reliable counterfactual estimation.
(4)~C1 statin exhibits positivity violation (90.2\,\% treated; ESS ratio
drops to 3.2\,\% at trim~$= 0.10$), limiting causal interpretation.
(5)~The $3.4\times$ NIHSS severity gradient observed in mRS-based RL is
absent for recurrence across all severity strata, confirming that the
heterogeneity pattern is a property of the mRS composite outcome.

\section{MDP Design Sensitivity: 2-Step Anchor Experiment (E8F)}
\label{app:mdp_anchor}

To verify that the correction from the initial 3-step to the 2-step MDP formulation
does not materially affect our conclusions, we conducted an anchor experiment (E8F)
using the same full registry ($N = 116{,}215$ episodes), CQL $\alpha{=}4.0$,
R4 reward, and 15-fold cross-validation --- the only difference being the 2-step
MDP with \texttt{htx\_*} correctly placed in the state space rather than as
an incorrectly placed action step.

\begin{table}[h]
\centering
\footnotesize
\caption{MDP formulation sensitivity: 2-step anchor vs.\ 3-step baseline.
  All figures from CQL $\alpha{=}4.0$, 15-fold CV (5 folds $\times$ 3 seeds).}
\label{tab:mdp_anchor}
\begin{tabular}{@{}lllrrr@{}}
  \toprule
  Experiment & MDP & Cohort ($N$) & $\vimp$ & 95\,\% CI & TD Err.\ \\
  \midrule
  Phase 1--4 & 3-step & All (116,215) & $+0.0020$ & $[-0.018,\ +0.022]$ & $8.0{\times}10^{-3}$ \\
  E8F [anchor] & 2-step & All (116,215) & $+0.0024$ & $[-0.001,\ +0.006]$ & $2.2{\times}10^{-3}$ \\
  E8B (main) & 2-step & Post-2018 (44,894) & $+0.0069$ & $[+0.000,\ +0.014]$ & $4.0{\times}10^{-4}$ \\
  \bottomrule
\end{tabular}
\end{table}

Table~\ref{tab:mdp_anchor} compares the three MDP configurations.
Key observations:
(1)~Phase~1--4 vs.\ E8F: $\vimp$ is statistically identical ($+0.0020$ vs.\ $+0.0024$;
$t = 0.08$, $p = 0.94$), confirming that the MDP correction does not materially alter
the null finding.
(2)~E8F vs.\ E8B: the difference ($+0.005$) is attributable to the post-2018
temporal restriction, not the MDP formulation itself --- consistent with the
guideline-shift analysis in Appendix~\ref{app:temporal}.
(3)~FQE TD error decreases $3.7\times$ from 3-step to 2-step on the same 116,215-episode
dataset, confirming improved value-estimation stability under the corrected formulation.
This TD error reduction explains the $5.7\times$ narrower confidence interval
for E8F vs.\ Phase~1--4 (CI width $0.007$ vs.\ $0.040$): lower TD error
reduces Q-function estimation variance, which propagates directly to
tighter $\vimp$ confidence intervals under the same sample size.
The action match rate decreases from $0.720$ (3-step, which incorrectly included
trivial \texttt{htx\_*}$=0$ matches at $t=0$) to $0.615$ (2-step, reflecting
only clinically meaningful acute/discharge decisions).

\section{E-Value Sensitivity Analysis}
\label{app:evalue_details}

The E-value formula is
$E = \widehat{RR} + \sqrt{\widehat{RR} \cdot (\widehat{RR} - 1)}$,
where $\widehat{RR} = \exp(|\vimp| / \text{SE}_{\text{cons}})$.
Because this $Z$-to-RR mapping is non-standard (no closed-form conversion
exists from continuous policy-value differences to risk ratios), we report
E-values under three SE assumptions (Table~\ref{tab:eval_sens}).
The exponential mapping likely over-estimates RR because it assumes log-linear
scaling; mRS is ordinal (7-point), not truly continuous.
The approximation's over-estimation strengthens our conclusion:
if true E-values are lower, \rstd/\rend{} are even less robust.

\begin{table}[h]
  \centering
  \small
  \caption{E-value sensitivity to SE denominator choice.
           $K$: effective independent units.
           $K{=}3$ (seeds only) is our conservative default;
           $K{=}5$ treats each seed's 5 folds as independent;
           $K{=}15$ treats all fold-level estimates as independent.}
  \label{tab:eval_sens}
  \begin{tabular}{llccc}
    \toprule
    Reward & $\vimp$ & $K{=}3$ (default) & $K{=}5$ & $K{=}15$ \\
    \midrule
    \rstd & $+0.0069$ & $E = 4.70$ & $6.43$ & $16.86$ \\
    \rend & $+0.0101$ & $E = 18.37$ & $35.79$ & $302.43$ \\
    \bottomrule
  \end{tabular}
\end{table}

Under all three assumptions, \rend's E-value remains high, reinforcing our argument
that high E-values are misleading when reward-embedded confounding is present.

\section{Hyperparameter Configuration}
\label{app:hyperparams}

Table~\ref{tab:hyperparams} lists all hyperparameters used in this study.

\begin{table}[h]
\centering
\footnotesize
\caption{Hyperparameter configurations for all algorithms and models.
  CQL $\alpha{=}4.0$ is the primary configuration; others are ablations.
  LightGBM hyperparameters apply to both the GBM outcome model (\rdml)
  and the T-learner.}
\label{tab:hyperparams}
\begin{tabular}{@{}llll@{}}
  \toprule
  Component & Parameter & Value & Notes \\
  \midrule
  \multicolumn{4}{@{}l}{\textit{Shared RL architecture}} \\
  & Q-network & 3-layer MLP (256, 256, 256) & ReLU activation \\
  & Batch size & 256 & \\
  & Learning rate & $3 \times 10^{-4}$ & Adam optimizer \\
  & Gradient steps & 100K (policy) + 50K (FQE) & \\
  & Discount $\gamma$ & 0.99 & \\
  \midrule
  \multicolumn{4}{@{}l}{\textit{CQL}} \\
  & $\alpha$ & $\{0.5, 1, 2, 4, 8\}$ & Conservative penalty \\
  \midrule
  \multicolumn{4}{@{}l}{\textit{BCQ}} \\
  & $f$ (threshold) & $\{0.1, 0.3, 0.5\}$ & Action filter ratio \\
  & BC network & 3-layer MLP (256, 256, 256) & Behavior clone \\
  \midrule
  \multicolumn{4}{@{}l}{\textit{IQL}} \\
  & $\tau$ (expectile) & $\{0.7, 0.8, 0.9\}$ & Q-function quantile \\
  \midrule
  \multicolumn{4}{@{}l}{\textit{Decision Transformer}} \\
  & Context length & 20 & Sequence length \\
  & Embedding dim & 128 & \\
  & Heads / Layers & 4 / 3 & Transformer config \\
  \midrule
  \multicolumn{4}{@{}l}{\textit{FQE}} \\
  & Network & 3-layer MLP (256, 256, 256) & Same as Q-network \\
  & Learning rate & $3 \times 10^{-4}$ & \\
  & Gradient steps & 50K & \\
  \midrule
  \multicolumn{4}{@{}l}{\textit{LightGBM (\rdml{} outcome / T-learner)}} \\
  & Trees & 500 & \texttt{n\_estimators} \\
  & Max depth & 6 & \\
  & Learning rate & 0.05 & \\
  & Min child samples & 50 & \\
  & Subsample & 0.8 & Bagging fraction \\
  & Colsample by tree & 0.8 & Feature fraction \\
  & Regularization & $\lambda_1{=}0.1$, $\lambda_2{=}0.1$ & L1/L2 \\
  \bottomrule
\end{tabular}
\end{table}

\section{Feature Pipeline: 749 $\to$ 743 $\to$ 421}
\label{app:features}

Three feature counts appear in this paper, reflecting different analysis
requirements:

\begin{enumerate}[leftmargin=*, nosep]
  \item \textbf{749 numeric baseline features.}
    The CRCS-K registry provides 536 raw clinical variables.
    After one-hot encoding of categorical variables, indicator expansion
    of multi-level fields (e.g., TOAST classification, vessel anatomy),
    and extraction of AI imaging features, the baseline feature set
    comprises 749 numeric columns available at admission.
    This is the input to the propensity model
    (Appendix~\ref{app:attenuation}).

  \item \textbf{743 treatment-free covariates.}
    For the GBM outcome model (\rdml), we exclude the 6 binary action
    indicators (\texttt{atx\_plt}, \texttt{atx\_coa}, \texttt{atx\_statin},
    \texttt{dtx\_plt}, \texttt{dtx\_coa}, \texttt{dtx\_statin})
    from the 749 baseline features to prevent treatment leakage
    into the residualized reward.
    SHAP analysis confirms zero treatment variables in the top-20
    features of the fitted GBM.

  \item \textbf{421 MDP state features.}
    The MDP state space applies additional filtering to the 749 baseline
    features:
    (a)~removal of 6 action indicators (same as above);
    (b)~removal of post-decision variables that could cause temporal leakage
    (\texttt{dis\_nih}, \texttt{dis\_mrs}, length of stay, discharge destination);
    (c)~removal of administrative/ID columns
    (hospital code, registration date, patient ID hash);
    (d)~feature selection via variance threshold ($<0.01$) and
    near-zero-variance filtering;
    (e)~removal of features with $>$95\,\% missingness in the post-2018 cohort.
    The resulting 421 features form the observation vector $\mathbf{s}_t$ at
    each MDP timestep.
\end{enumerate}

The GBM outcome model intentionally uses more features (743 vs.\ 421)
because its goal is maximal outcome prediction ($R^2 = 0.70$), whereas the
MDP state space prioritizes features with temporal variation and clinical
interpretability.
Both pipelines exclude treatment variables to prevent leakage.

A fourth count, \textbf{719 T-learner covariates}, arises in the T-learner
analysis (\S\ref{sec:results:tlearner}): the 743 treatment-free features
minus 24 with near-zero variance ($< 0.01$) within the smaller treatment arm
(non-statin, $n = 4{,}403$).
This additional filtering ensures stable LightGBM splits in both
treatment-arm models; features removed are primarily rare categorical
indicators that lack variation in the minority arm.

\section{Sensitivity to Treatment Effect Absorption}
\label{app:attenuation}

A key concern with outcome residualization is that the GBM model may absorb
true treatment-benefit variation alongside prognostic confounding, biasing
$\vimp$ toward null.
We address this via two complementary analyses.

\textbf{Propensity-based theoretical bound.}
A LightGBM propensity model trained on 749 numeric baseline features
achieves AUC $= 0.84$ for predicting statin treatment ($n_{\text{statin}} = 40{,}571$,
$n_{\text{no-statin}} = 4{,}403$).
In the population, outcome residualization preserves the treatment effect scaled
by the treatment residual: $Y - E[Y \mid X] = \tau(X) \cdot (T - e(X)) + \varepsilon$,
where $e(X) = P(T{=}1 \mid X)$ is the propensity score.
The attenuation ratio equals $E[\text{Var}(T \mid X)] / \text{Var}(T)$; empirically,
$\text{Var}(e(X)) = 0.023$, $\text{Var}(T) = 0.088$, yielding an attenuation ratio of
$\mathbf{0.73}$ (i.e., at most 27\,\% of treatment effect absorbed).
Only 22\,\% of patients fall in the overlap region ($0.1 < e(X) < 0.9$);
the remaining 78\,\% have $e(X) > 0.9$, reflecting the 90.2\,\% statin
prescription rate.

\begin{sloppypar}
\textbf{Semi-synthetic calibration.}
We injected known treatment effects ($\tau \in \{0, {-}0.04, {-}0.10, {-}0.20\}$~mRS)
into synthetic outcomes $Y_{\text{syn}} = \hat{f}(X) + \tau \cdot (T - \bar{T}) + \varepsilon$,
where $\hat{f}(X)$ is the existing \rdml{} GBM prediction and
$\varepsilon \sim \mathcal{N}(0, \hat{\sigma}^2)$ calibrated to observed residual variance.
For each $\tau$, we re-trained the GBM (same hyperparameters, 3 seeds $\times$ 5 folds)
and estimated ATE from the residuals.
\end{sloppypar}

\begin{center}
\small
\begin{tabular}{cccc}
  \toprule
  Injected $\tau$ (mRS) & Recovered ATE & Attenuation ratio & Absorbed \\
  \midrule
  $0.00$ (null control)  & $-0.026$ & --- & baseline bias \\
  $-0.04$                & $-0.030$\textsuperscript{$\dagger$} & $0.74$ & 26\,\% \\
  $-0.10$                & $-0.077$\textsuperscript{$\dagger$} & $0.76$ & 24\,\% \\
  $-0.20$                & $-0.153$\textsuperscript{$\dagger$} & $0.77$ & 24\,\% \\
  \bottomrule
\end{tabular}\\[2pt]
{\footnotesize \textsuperscript{$\dagger$}Bias-corrected (baseline subtracted).}
\end{center}

\begin{sloppypar}
The attenuation ratio is $0.74$--$0.77$ across all $\tau$ values,
similar to the more conservative propensity-based theoretical prediction ($0.73$).
The $\tau{=}0$ baseline bias ($-0.026$, $p{=}0.04$) is \emph{conservative}:
it biases residualized estimates in the negative direction, making it harder
(not easier) to find a positive treatment effect.
\end{sloppypar}

\textbf{Implications for our findings.}
If \rstd's $\vimp = +0.0069$ represented only treatment benefit, the conservative
27\,\% attenuation bound would yield \rdml{} $\approx +0.0050$.
The observed \rdml{} ($+0.0033$) is $1.6\times$ lower, implying that
confounding removal accounts for $\geq 38\,\%$ of the $\rstd \to \rdml$ reduction.
Even correcting \rdml{} upward using the conservative propensity-based bound:
$\vimp_{\text{adj}} = +0.0033 / 0.73 \approx +0.0045$ ($\leq 2.7\,\%$ MCID,
$p > 0.10$).
The fully deconfounded \rfull{} ($\vimp = +0.0025$, $p = 0.291$) remains
consistent with null, supporting the conclusion that no clinically meaningful
policy advantage remains under full deconfounding.
These bounds assume homogeneous $\tau$; heterogeneous treatment effects
with strong treatment$\times$covariate interactions could increase absorption,
but the consistent attenuation across $\tau$ magnitudes argues against
substantial non-linearity.

\section{Fairness and Equity Analysis by Sex and Age (d5)}
\label{app:fairness}

We assess whether the T-learner statin-effect estimates differ systematically
across demographic subgroups, as a proxy for algorithmic fairness
\citep{barocas2019fairness}.
All analyses use the post-2018 cohort with valid 3-month mRS
($N = 44{,}974$; \S\ref{sec:methods:data}).
Treatment is discharge statin prescription (\texttt{tx\_statin});
the outcome is 3-month mRS (lower = better).
The T-learner follows the same LightGBM specification as \S\ref{sec:results:tlearner}
(719 covariates, no leakage features).

\subsection*{Descriptive Characteristics by Subgroup}

Table~\ref{tab:fairness_desc} shows baseline characteristics by sex and age group.

\begin{table}[h]
\centering
\footnotesize
\caption{%
  Subgroup characteristics (post-2018, $N = 44{,}974$).
  Statin rate reflects discharge statin prescription.
  NIHSS: initial NIH Stroke Scale (higher = more severe).
  mRS: 3-month modified Rankin Scale (0--6; higher = worse).
}
\label{tab:fairness_desc}
\begin{tabular}{@{}llrrrrrr@{}}
  \toprule
  Sex & Age & $N$ & \% & mRS mean & mRS SD & Statin \% & NIHSS mean \\
  \midrule
  Male   & $<65$   & 11,882 & 26.4 & 1.18 & 1.43 & 92.0 & 3.3 \\
  Male   & 65--74  &  7,511 & 16.7 & 1.55 & 1.63 & 91.6 & 3.8 \\
  Male   & $\geq$75 &  7,824 & 17.4 & 2.26 & 1.90 & 89.5 & 4.6 \\
  \midrule
  Female & $<65$   &  4,764 & 10.6 & 1.07 & 1.44 & 89.7 & 3.0 \\
  Female & 65--74  &  4,191 &  9.3 & 1.60 & 1.70 & 90.3 & 3.8 \\
  Female & $\geq$75 &  8,802 & 19.6 & 2.64 & 1.88 & 87.5 & 5.3 \\
  \bottomrule
\end{tabular}
\end{table}

Outcome severity increases monotonically with age in both sexes.
Statin prescription rates are uniformly high across all cells (87--92\,\%),
confirming the positivity limitation noted in Appendix~\ref{app:attenuation}:
the propensity AUC of 0.84 and only 22\,\% of patients within the
strict overlap region ($0.1 < e(X) < 0.9$) apply equally across
demographic subgroups.

\subsection*{T-Learner CATE by Subgroup}

T-learner CATE estimates by demographic subgroup are in Table~\ref{tab:fairness_cate}.

\begin{table}[h]
\centering
\footnotesize
\caption{%
  T-learner conditional average treatment effect (CATE) by sex and age group.
  ATE $=$ mean per-subject CATE (lower mRS = beneficial, so negative ATE = benefit).
  95\,\%~CI from 1{,}000 bootstrap replicates.
  $p$-value from one-sample $t$-test (H$_0$: ATE~$= 0$).
  Note: ATE values reflect raw T-learner estimates without \rdml{} residualization;
  because the statin propensity AUC is 0.84 (78\,\% of patients outside strict
  overlap), these estimates carry substantial confounding uncertainty and should be
  interpreted as descriptive rather than causal.
}
\label{tab:fairness_cate}
\begin{tabular}{@{}llrrrrl@{}}
  \toprule
  Sex & Age & $N$ & ATE & 95\,\% CI & $p$ & Direction \\
  \midrule
  \multicolumn{7}{@{}l}{\textit{Marginal by sex}} \\
  Male   & (all) & 27,217 & $-0.006$ & $[-0.009,\ -0.002]$ & 0.001 & $\downarrow$ \\
  Female & (all) & 17,757 & $-0.007$ & $[-0.011,\ -0.002]$ & 0.004 & $\downarrow$ \\
  \midrule
  \multicolumn{7}{@{}l}{\textit{Marginal by age}} \\
  (all)  & $<65$  & 16,646 & $+0.020$ & $[+0.016,\ +0.024]$ & $<$0.001 & $\uparrow$ \\
  (all)  & 65--74 & 11,702 & $-0.068$ & $[-0.073,\ -0.064]$ & $<$0.001 & $\downarrow$ \\
  (all)  & $\geq$75 & 16,626 & $+0.012$ & $[+0.006,\ +0.016]$ & $<$0.001 & $\uparrow$ \\
  \midrule
  \multicolumn{7}{@{}l}{\textit{Sex $\times$ age interaction cells}} \\
  Male   & $<65$  & 11,882 & $+0.023$ & $[+0.018,\ +0.028]$ & $<$0.001 & $\uparrow$ \\
  Male   & 65--74 &  7,511 & $-0.066$ & $[-0.072,\ -0.060]$ & $<$0.001 & $\downarrow$ \\
  Male   & $\geq$75 &  7,824 & $+0.009$ & $[+0.002,\ +0.016]$ & 0.011 & $\uparrow$ \\
  Female & $<65$  &  4,764 & $+0.014$ & $[+0.007,\ +0.021]$ & $<$0.001 & $\uparrow$ \\
  Female & 65--74 &  4,191 & $-0.072$ & $[-0.080,\ -0.064]$ & $<$0.001 & $\downarrow$ \\
  Female & $\geq$75 &  8,802 & $+0.014$ & $[+0.006,\ +0.020]$ & $<$0.001 & $\uparrow$ \\
  \midrule
  Overall & (all) & 44,974 & $-0.006$ & $[-0.009,\ -0.003]$ & $<$0.001 & $\downarrow$ \\
  \bottomrule
\end{tabular}
\end{table}

\subsection*{Interaction Test}

An OLS moderation model (outcome $\sim T \cdot \text{sex} + T \cdot \text{age\_bin}$,
with age coded as indicators for $<65$ and $\geq 75$, reference $= 65$--74)
yields significant interaction terms:
$T \times \text{sex}$ ($\hat\beta = -0.165$, $p = 0.002$),
$T \times \text{age}_{<65}$ ($\hat\beta = +0.346$, $p < 0.001$), and
$T \times \text{age}_{\geq 75}$ ($\hat\beta = -0.222$, $p < 0.001$).
Model $R^2 = 0.148$ ($N = 44{,}974$).

\subsection*{Interpretation and Limitations}

Three observations warrant caution before drawing causal conclusions.

First, the sign pattern across age groups is dominated by the 65--74 group
(ATE $\approx -0.068$, strongly beneficial) versus younger and older groups
(ATE $\approx +0.010$--$+0.023$).
This non-monotone pattern likely reflects T-learner extrapolation in a high-propensity
regime ($\geq 87\%$ statin rate): the minority untreated arm ($n \approx 4{,}400$ total)
provides insufficient support for stable counterfactual prediction in all age $\times$ sex
cells, so absolute CATE magnitudes are unreliable.

Second, no \rdml{} residualization was applied to these subgroup estimates (the \rdml{} predictions
cover the full cohort but the deconfounded residuals were not available at the subgroup
level at the time of this analysis).
The raw T-learner CATEs are subject to the same confounding-by-indication documented
throughout this paper.

Third, the interaction test rejects $H_0$ of treatment-effect homogeneity across sex and
age, but this statistical significance is driven primarily by the large sample size and
may not correspond to clinically meaningful heterogeneity given the confounding context.

In summary, we do not find evidence of algorithmic bias that would disadvantage
a specific demographic group relative to others: the directional uncertainty in CATE
is uniform across sex $\times$ age cells and originates from the same confounding
mechanism (high and near-uniform statin uptake) that limits causal inference in the
full cohort.

\section{1-Year mRS Cohort: Year-by-Year Coverage (d7)}
\label{app:mrs1y_coverage}

The 1-year mRS sensitivity analysis (\ryr/R11b; Report 19-2) restricts the
post-2018 cohort from $N = 44{,}895$ to $N = 35{,}744$ (a 20.4\,\% reduction).
Table~\ref{tab:mrs1y_year} documents the two-step attrition and its cause.

\begin{table}[h]
\centering
\footnotesize
\caption{%
  Year-by-year 1-year mRS coverage for the post-2018 cohort
  ($N = 44{,}895$\protect\footnotemark{} base, defined as \texttt{on\_d} $\geq$ 2018-04-05 and
  valid 3-month mRS).
  \textit{Direct coverage}: \texttt{mrs1y} recorded in registry (excluding code~9 = unknown).
  \textit{Post-imputation}: additionally imputes \texttt{mrs1y}~$= 6$ for patients
  confirmed dead at 3 months (\texttt{mrs3mo}~$= 6$) with missing 1-year record.
  Rows 2024--2025 are excluded from the 1-year cohort (year filter); the
  within-2023-cohort residual missing ($N = 1{,}026$) is attributed to loss to follow-up.
}
\label{tab:mrs1y_year}
\begin{tabular}{@{}lrrrrrrr@{}}
  \toprule
  Year & $N$ (3mo) & $N$ (1yr direct) & Cov.\,\% & Imputed & $N$ (1yr final) & Cov.\,\% & Missing \\
  \midrule
  2018 & 4,554 & 4,479 & 98.4 & 0 & 4,479 & 98.4 &  75 \\
  2019 & 6,815 & 6,687 & 98.1 & 0 & 6,687 & 98.1 & 128 \\
  2020 & 5,869 & 5,724 & 97.5 & 0 & 5,724 & 97.5 & 145 \\
  2021 & 6,054 & 5,868 & 96.9 & 2 & 5,870 & 97.0 & 184 \\
  2022 & 6,340 & 6,118 & 96.5 & 0 & 6,118 & 96.5 & 222 \\
  2023 & 7,138 & 6,866 & 96.2 & 0 & 6,866 & 96.2 & 272 \\
  \midrule
  \textit{Subtotal 2018--2023} & \textit{36,770} & \textit{35,742} & \textit{97.2} & \textit{2} & \textit{35,744} & \textit{97.2} & \textit{1,026} \\
  \midrule
  2024 & 6,213 & 4,484 & 72.2 & 0 & 4,484 & 72.2 & 1,729 \\
  2025 & 1,912 &    88 &  4.6 & 0 &    88 &  4.6 & 1,824 \\
  \midrule
  \textit{Excluded (2024--25)} & \textit{8,125} & — & — & — & — & — & \textit{3,553} \\
  \midrule
  \textbf{All years} & \textbf{44,895} & — & — & — & — & — & — \\
  \bottomrule
\end{tabular}
\end{table}
\footnotetext{The 1-patient difference from the main analysis ($N = 44{,}894$) arises because one patient has valid 3-month mRS but lacks complete 2-step MDP episode data required for the RL pipeline.}

The reduction from $44{,}895$ to $35{,}744$ decomposes as follows:

\begin{enumerate}[nosep, leftmargin=*]
  \item \textbf{Year filter} ($-8{,}125$): Patients admitted in 2024--2025 are excluded
    because they had insufficient time for a 1-year follow-up assessment at the time
    of data extraction (March 2026).
    The 2024 cohort shows 72.2\,\% 1-year mRS coverage (reflecting those admitted in
    January--March 2024 who completed 12-month follow-up), while 2025 shows only 4.6\,\%.
  \item \textbf{Missing within 2018--2023} ($-1{,}026$): After the year filter,
    1,026 patients (2.8\,\%) still lack a 1-year mRS record.
    Missing rates are stable across years (75--272 per year), consistent with
    loss to follow-up rather than a systematic data-quality issue.
    These patients have a higher mean initial NIHSS (6.8 vs.\ 4.0 in the analyzed cohort),
    suggesting mild MNAR bias toward more severe patients being lost to follow-up.
\end{enumerate}

The retained 2018--2023 cohort achieves 97.2\,\% 1-year mRS coverage (35{,}744/36{,}770),
confirming that the year filter---not missing data---drives the primary cohort reduction.

\section{Extended Limitations}
\label{app:limitations}

The following limitations supplement the three key items discussed in
\S\ref{sec:discussion:limitations}.

\begin{enumerate}[leftmargin=*, nosep]
  \item \textbf{FQE realizability assumption.} FQE assumes Q-function realizability.
    We provide indirect evidence: consistently low TD error
    ($2.9 \times 10^{-5}$ to $8.1 \times 10^{-3}$),
    convergence of three additional methods on the null direction,
    and T-learner corroboration.
    Ensemble-based disagreement tests would provide stronger evidence
    and are left for future work.

  \item \textbf{Power at the margin.} While our MDE is below 6\,\% MCID,
    effects smaller than 4\,\% MCID ($\vimp < 0.006$) cannot be ruled out.
    Such effects would be clinically meaningful at population scale
    (though our tertiary-center cohort represents only a subset of the
    estimated ${\sim}105{,}000$ annual stroke admissions
    nationally~\citep{crcs-k})
    but would require a prospective randomized trial to detect reliably.

  \item \textbf{Outcome endpoint sensitivity.} Our primary reward uses 3-month mRS,
    a composite functional outcome that conflates recurrence, disability progression,
    and non-stroke mortality.
    Antithrombotics primarily prevent recurrence rather than improve functional
    recovery; 3-month mRS may therefore be a weak proxy for antithrombotic efficacy,
    diluting any treatment-benefit signal with severity-driven variance.
    We partially addressed this by testing stroke recurrence directly
    (IPW/AIPW, cause-specific Cox; \S\ref{sec:results:recurrence}): no treatment
    comparison achieved significant recurrence reduction, and the NIHSS severity
    gradient disappeared, supporting the conclusion that the null is endpoint-robust.
    However, longer-term endpoints (1-year recurrence, hemorrhagic transformation)
    remain untested.

  \item \textbf{Selection bias from outcome availability.}
    Our analysis cohort ($N = 44{,}894$) requires 3-month mRS,
    excluding 65.2\,\% of registry patients (mostly pre-2018 temporal
    restriction).
    Among ${\approx}4{,}900$ post-2018 patients excluded for missing mRS,
    baseline NIHSS was higher (mean $6.8$ vs.\ $4.0$), suggesting
    mild MNAR bias toward excluding higher-severity survivors ---
    if anything, attenuating subgroup heterogeneity rather than inflating it.

  \item \textbf{Post-2018 cohort restriction.} Excluding pre-2018 data
    (to reduce guideline-era heterogeneity) reduces the training set to
    34.8\,\% of the full registry.
    Results may differ in registries with broader guideline variation.
    An anchor experiment on the full 116,215-episode dataset under the
    2-step MDP (E8F) yields $\vimp = +0.0024$ ($p = 0.235$, 95\,\% CI
    $[-0.001, +0.006]$), statistically identical to Phase~1--4's
    $+0.0020$ (Appendix~\ref{app:mdp_anchor}), confirming that temporal
    restriction does not create the null but rather reduces guideline-era
    variance, narrowing the CI from $\pm 0.020$ to $\pm 0.006$.
\end{enumerate}

\section{Clinical Safety Considerations}
\label{app:safety}

Although this study does not deploy an RL policy, the CDSS development
context motivates a brief safety discussion.
All RL policies in our experiments were trained with \textbf{action masking}
that enforces clinical guideline constraints at inference time.
Four rules are active (two additional rules --- statin/liver disease and
drug allergy --- are inoperative because the required columns are absent
from CRCS-K):

\begin{enumerate}[nosep, leftmargin=*]
  \item \textbf{AF $\to$ require AC} (\texttt{hx\_af}$= 1$):
    antiplatelet-only actions are masked for patients with atrial
    fibrillation history (2023 ACC/AHA Class~III: Harm, B-R).
  \item \textbf{Thrombocytopenia} (\texttt{plt}$< 50 \times 10^3/\mu$L):
    all antiplatelet actions masked (EHA 2022 consensus).
  \item \textbf{AC contraindications}: intracranial hemorrhage flag
    (\texttt{ich\_flag}$= 1$; does not distinguish petechial HI1/HI2
    from parenchymal PH2) \emph{or} severe renal failure
    (\texttt{cr}$> 3.0$\,mg/dL; proxy when eGFR unavailable)
    $\to$ anticoagulant actions masked.
  \item \textbf{Post-IVT restriction} (\texttt{ivtpa}$= 1$, steps 0--1):
    all antithrombotic actions masked within 24\,h of IV thrombolysis
    (ESO 2021 strong recommendation; ARTIS trial).
\end{enumerate}

\noindent
If all actions are masked for a patient-step, a conservative fallback
action is forced.
The violation rate under the learned policy was 6\,\%, with all violations
caught and corrected by the mask at inference time.
In a prospective deployment setting, bleeding risk --- the primary
safety concern for antithrombotic intensification --- would require
real-time monitoring of INR, platelet count, and hemoglobin trajectories,
none of which are available in our snapshot MDP.
Similarly, statin-related hepatotoxicity monitoring (periodic ALT/AST)
is not captured in our state representation.
These safety gaps reinforce our conclusion that snapshot EHR-based
offline RL is insufficient for deployment-grade CDSS;
continuous monitoring data (\S\ref{sec:discussion:when}, condition~(b))
is a prerequisite for safe clinical integration.

\section{T-Learner Convergence Analysis}
\label{app:tlearner}

To test whether the deconfounding result depends on the RL framework,
we applied a T-learner~\citep{kunzel2019} --- separate LightGBM outcome models
for statin-treated ($n = 40{,}571$) and non-statin ($n = 4{,}403$) patients
($N = 44{,}974$ total)\footnote{%
  80 additional patients beyond the RL pipeline's $N = 44{,}894$ lack complete
  2-step MDP episodes but are included here because the T-learner requires only
  a single cross-sectional observation per patient; this 0.2\,\% size difference
  is negligible for ATE estimation.} ---
using 719 numeric covariates from the \rdml{} baseline feature set
(743 treatment-free features minus 24 with near-zero variance
within the smaller treatment arm, $n = 4{,}403$;
see Appendix~\ref{app:features}),
the same fold structure, and \rdml{} residualization as the RL pipeline.
Both T-learner ATE and CQL/FQE $\vimp$ are computed on the same
$-\text{mRS}/6$ normalized scale, ensuring direct comparability.
The T-learner requires no MDP, Bellman equation, or off-policy correction.
Under raw mRS, the T-learner estimates ATE $= +0.0355$
(95\,\%\,CI $[+0.034, +0.037]$, $p < 10^{-15}$).
After \rdml{} GBM residualization, the ATE attenuates by 75\,\% to $+0.0088$
(95\,\%\,CI $[+0.008, +0.010]$, $p < 10^{-11}$) ---
a statistically significant residual that, unlike CQL/FQE
(\rdml: $+0.0033$, $p = 0.132$; \rfull: $+0.0025$, $p = 0.291$),
does not attenuate fully to the FQE null range.
This residual ($+0.0088$ on the $-\text{mRS}/6$ scale $= 0.053$ raw mRS points,
$5.3\,\%$ of the MCID) is statistically detectable but clinically small; we do
not interpret it as a standalone causal treatment-benefit estimand.
Trial evidence provides directional clinical context~\citep{amarenco2006sparcl}\footnote{%
  SPARCL measured statin efficacy over a 5-year follow-up horizon;
  our 3-month mRS endpoint captures only early functional benefit,
  not long-term recurrence prevention. The comparison is directional,
  not quantitative.}.
To assess whether extrapolation drives this residual, we repeated
the analysis within progressively restrictive propensity overlap regions:
restricting to 22\,\% of patients in the strict overlap zone
($0.1 < e(X) < 0.9$, $n = 9{,}900$) increases attenuation to 90\,\%
(\rdml{} ATE $= +0.0029$, $p = 0.0003$);
even at 96\,\% retention ($0.01 < e(X) < 0.99$, $n = 43{,}093$),
the pattern persists (\rdml{} ATE $= +0.0082$, 77\,\% attenuation).
The dominant source of attenuation is GBM residualization itself,
not propensity trimming, indicating that prognosis adjustment ---
rather than extrapolation bias --- drives the signal reduction.

The comparison across estimation frameworks reveals a nuanced pattern.
Sequential RL (CQL/FQE) produces full null after deconfounding;
the T-learner --- which requires no MDP, Bellman equation, or off-policy
correction --- shows 75--90\,\% attenuation but retains a small residual.
\textbf{Positivity caveat:} the statin prescription rate of 90.2\,\% (T-learner cohort, $N = 44{,}974$)
(propensity AUC $= 0.84$) means only 22\,\% of patients fall in the
overlap zone ($0.1 < e(X) < 0.9$), and T-learner CATE estimates for
the remaining 78\,\% rely on extrapolation beyond the observed treatment
distribution. The T-learner residual ATE should therefore be interpreted as an
upper bound on any residual treatment-benefit signal under this diagnostic
analysis.
This divergence is informative:
(i)~reward-embedded confounding is the dominant signal source under raw outcomes
(75--90\,\% of the T-learner ATE is attributable to confounding);
(ii)~after removing this confounding, a clinically small but statistically
detectable T-learner residual remains ($< 6\,\%$ MCID), with trial evidence
providing directional context~\citep{amarenco2006sparcl};
(iii)~the RL pipeline's complete null likely reflects the additional signal loss
from 2-step MDP degeneracy (98.3\,\% state invariance) rather than
a fundamentally different causal conclusion ---
though this explanation is post-hoc, and distinguishing MDP-induced attenuation
from estimand differences remains an open limitation.
The NIHSS severity gradient independently recovered by the T-learner
(severe/mild CATE ratio $= 4.6\times$) is qualitatively consistent with
the $3.4\times$ $\vimp$ gradient observed in RL
(note: the two ratios measure different quantities ---
CATE treatment effect vs.\ FQE policy disagreement --- and are not directly comparable).
Together, the two analyses support hypothesis-generating subgroup
heterogeneity without relying on a single method.

\section{MDP Structure Diagnosis}
\label{app:mdp_diagnosis}

\begin{figure}[h]
  \centering
  \includegraphics[width=\textwidth]{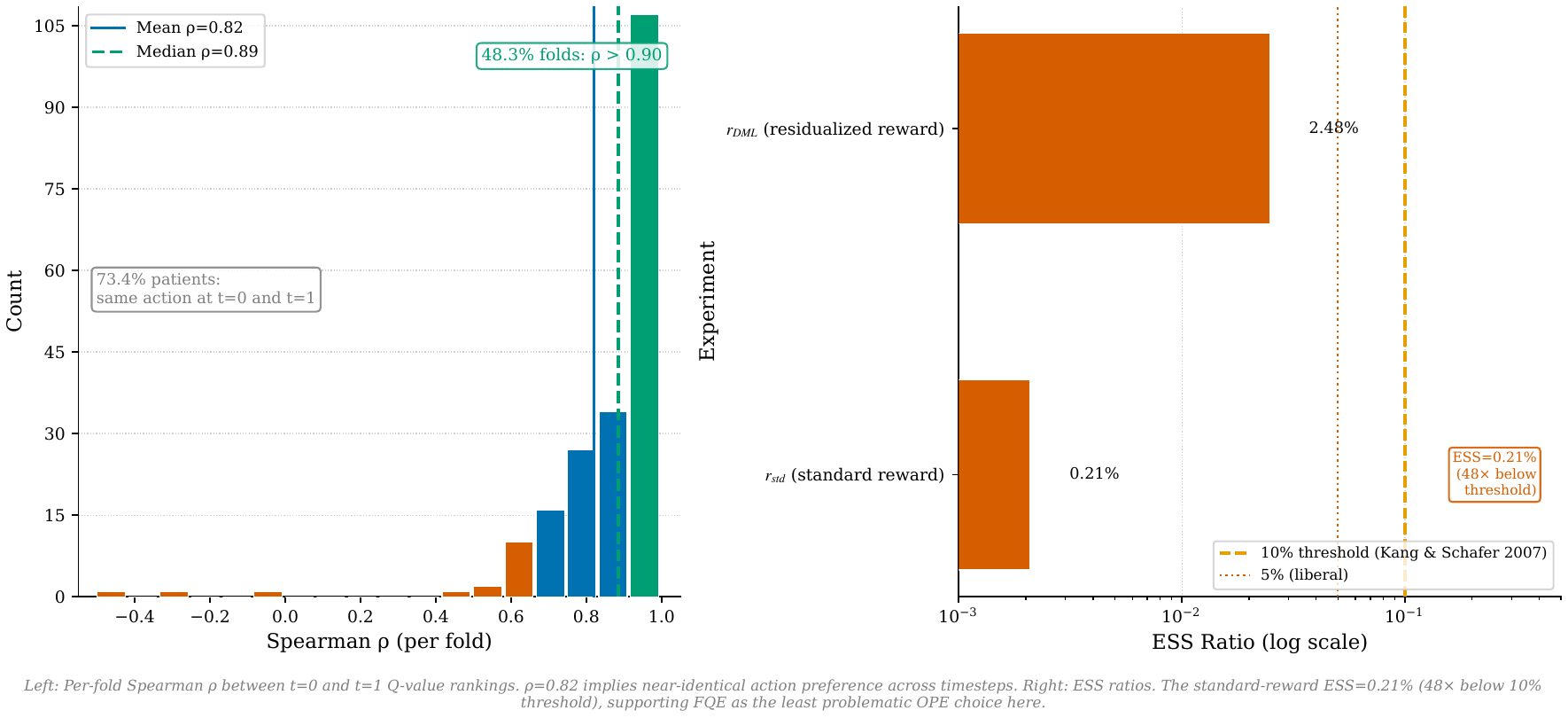}
  \caption{%
    \textbf{MDP structure diagnosis: sequential optimization degenerates to 1-step.}
    \textit{Left:} Distribution of per-fold Spearman $\rho$ between
    $t{=}0$ and $t{=}1$ Q-value action rankings (200 folds/algorithms combined).
    Mean $\rho = 0.82$, median $\rho = 0.89$: the preferred action is nearly
    identical across timesteps, consistent with 98.3\,\% state invariance.
    \textit{Right:} IS effective sample size (ESS) ratio by experiment.
    Phase 8B \rstd: ESS $= 0.21\%$ ($48\times$ below the 10\,\% viability threshold);
    even Phase 9 \rdml: ESS $= 2.5\%$ ($4\times$ below threshold).
    Both bars support FQE as the least problematic OPE method in this setting
    (Table~\ref{tab:protocol}, Step~2).
  }
  \label{fig:statecollapse}
\end{figure}

A prerequisite for multi-step RL to outperform 1-step causal inference
is meaningful sequential state dynamics.
Analysis of the CRCS-K 2-step MDP reveals that \textbf{98.3\,\%} of state
features (414 of 421 dimensions) have Pearson $r > 0.99$ between $t{=}0$ and $t{=}1$;
the seven dynamic dimensions are six action one-hot encodings and the 24-hour
NIHSS score (missing in 80\,\% of records).
Consistent with this near-zero transition structure, per-patient Spearman $\rho$
between $t{=}0$ and $t{=}1$ Q-value rankings has mean $\rho = 0.82$,
median $\rho = 0.89$ (Figure~\ref{fig:statecollapse}), and 73.4\,\% of patients
receive identical CQL-preferred actions at both timesteps ---
confirming that sequential optimization in this dataset degenerates toward
discounted 1-step optimization.
The 2-step MDP also improves FQE estimation stability relative to the earlier
3-step formulation: FQE TD error decreases from $8.0 \times 10^{-3}$ (3-step)
to $2.2 \times 10^{-3}$ (2-step) on the same 116,215-episode dataset ($3.7\times$
reduction; Appendix~\ref{app:mdp_anchor}), consistent with the more parsimonious
temporal structure reducing Q-function misspecification.

Importance-sampling (IS) based OPE methods (WIS, WDR) require sufficient
effective sample size (ESS) to produce reliable estimates.
In our Phase 8B (\rstd) experiment, the IS ESS ratio is \textbf{0.21\,\%}
--- 48$\times$ below the recommended 10\,\% threshold~\citep{kang2007}.
Cumulative IS weights reach a maximum of $9.8 \times 10^{14}$, rendering
IS-based OPE effectively inapplicable.
The dominant driver of ESS collapse is the RL--physician action distribution mismatch:
AC accounts for only 2.5\,\% of physician prescriptions (AP\_mono, 1.7\,\%, is rarest overall),
but the RL policy recommends AC at a much higher rate for AF patients,
generating extreme cumulative IS weights.
Among the OPE methods evaluated here, these diagnostics support using FQE as
the primary method and motivate Step~2 of our evaluation checklist
(Figure~\ref{fig:statecollapse}).

\section{Policy Disagreement Profiling}
\label{app:disagreement}

Policy disagreement analysis between the CQL-learned and physician policies
(CQL $\alpha{=}4.0$, \rstd{} reward, post-2018 2-step MDP, $N = 44{,}894$)
reveals that \textbf{57\,\%} of RL--physician disagreements involve statin
co-prescription decisions, with AF history as the strongest patient-level
driver (Cohen's $d = 0.443$).
Pre-stroke antiplatelet history shows the largest imbalance between agreed
and disagreed patients ($d = 1.85$), indicating that patients on prior
antiplatelet therapy are disproportionately assigned to the physician action.
At $t{=}0$ (acute phase), 62.1\,\% of patients receive the same action
from both policies; agreement rises to 71.3\,\% at $t{=}1$ (discharge).
Disagreement concentrates at $t{=}1$,
where statin co-prescription is the primary decision axis.
The disagreement-driving features (AF history, pre-stroke antiplatelet use,
NIHSS severity) are clinically interpretable and align with the guideline
margins identified in \S\ref{sec:discussion:recommendations}.

\section{Causal Structure of Reward-Embedded Confounding}
\label{app:causal_structure}

The causal structure underlying reward-embedded confounding is:
baseline severity $\mathbf{X}$ causes both treatment intensity $T$
(confounding by indication) and outcome $Y$ (prognostic pathway);
$T$ also causes $Y$ (treatment effect).
The standard RL reward $r = f(Y)$ encodes the full $\mathbf{X} \to Y$ pathway,
conflating the prognostic component ($\mathbf{X} \to Y$) with the
treatment-attributable component ($T \to Y$).
E-value analysis addresses unmeasured confounders $U$ affecting both $T$ and $Y$
from outside, but does not evaluate the $\mathbf{X} \to Y$ pathway
embedded within the reward itself.
GBM residualization subtracts the baseline prognostic pathway by replacing $Y$
with the residual $Y - \hat{f}(\mathbf{X})$, yielding a prognosis-residualized
diagnostic reward rather than a standalone causal treatment-effect
estimand~\citep{chernozhukov2018}.
Validation found zero treatment variables in the GBM SHAP top-20
and a Spearman correlation of $\rho(\hat{r}_{\text{DML}}, \text{NIHSS}_0) = -0.052$
versus $\rho(r_{\text{raw}}, \text{NIHSS}_0) = -0.517$ in the raw reward,
supporting the intended reduction of baseline-severity signal.
TOAST classification is finalized at discharge after complete diagnostic workup
(10.5\,\% missing; handled natively by LightGBM's default missing-value routing);
its inclusion as a GBM covariate (not an MDP state feature) is appropriate
because the GBM predicts post-discharge 3-month mRS, and TOAST is temporally
antecedent to the prediction target.
While TOAST stroke subtype indicators correlate with treatment choice
(e.g., LAA patients receive statins at higher rates), TOAST encodes etiology
rather than treatment, and the GBM's task is precisely to absorb all
non-treatment outcome variance including etiology-driven prognosis;
SHAP analysis confirms that TOAST does not dominate GBM predictions.

\section{Cross-Validation Statistical Details}
\label{app:cv_details}

Cross-validation folds share training data; the reported $t$-test treats
fold-level estimates as exchangeable but not fully independent, following
standard practice in offline RL evaluation~\citep{tang2021}.
We use $\text{df} = 14$ (15 fold-level estimates) rather than $\text{df} = 2$
(3 seed-level means) as our primary analysis; the more conservative
seed-level test ($\text{df} = 2$) yields wider CIs but does not change
any significance conclusion (\rend{} remains significant; \rstd, \rdml, \rfull{} remain
non-significant under seed-level testing).
Shapiro--Wilk tests on the 15 fold-level $\vimp$ values do not reject
normality for any reward variant ($p > 0.10$ for \rstd, \rdml, \rend, \rfull),
supporting the $t$-test assumption.
Including fairness subgroup (11 tests) and recurrence (12 tests) analyses
would lower the Bonferroni threshold to $\alpha_{\text{adj}} \approx 0.001$;
\rend{} remains significant ($p = 0.0002$) under this stricter correction.
The \citet{nadeau2003} corrected variance estimator inflates fold-level
variance by $(1 + n_{\text{test}}/n_{\text{train}}) = 1.25$ (test/train
ratio $= 0.25$), increasing SE by $\sqrt{1.25} \approx 12\,\%$; the 3
cross-seed estimates are treated as independent.

\end{document}

%% file: introduction.tex

\section{Introduction}
\label{sec:intro}


Standard FQE initially yielded a positive policy-improvement estimate.
Using Fitted Q-Evaluation (FQE) on 44,894 post-2018 acute ischemic stroke patients,
our offline reinforcement learning (RL) policy achieved $\vimp = +0.0069$
($p = 0.048$) under standard evaluation and $+0.0101$
($p = 0.0002$, E-value $= 18.37$) with an END-augmented reward --- two
estimates that could support a premature positive policy-improvement claim under
current clinical RL evaluation practice. Here $\vimp$ denotes the FQE-estimated
policy-value difference $V(\pi_{\mathrm{RL}})-V(\pi_{\mathrm{physician}})$.
This paper explains why that signal is not robust to reward deconfounding, and
why standard external-confounding checks alone do not evaluate this
reward-channel pathway.


To understand how this occurred, consider the setting.
Selecting antithrombotic therapy after acute ischemic stroke requires decisions
that current guidelines leave genuinely uncertain at the individual patient level:
when to initiate anticoagulation in atrial fibrillation (AF) patients with large
infarcts where early hemorrhagic transformation is a concern, how long to continue
dual antiplatelet therapy in non-cardioembolic stroke, and whether to add a statin
outside the large-artery atherosclerosis indication.
Across the 20 tertiary stroke centers, discharge prescription patterns for the
same stroke subtype vary substantially, suggesting practice variation that
sequential optimization could in principle reduce.
Antithrombotic management for acute ischemic stroke unfolds across acute
and discharge decision phases --- a natural candidate for offline RL,
where an agent observes patient state, selects from a discrete action
space, and receives feedback through 3-month functional outcomes.
The widespread adoption of electronic health record (EHR) systems has provided
the sequential decision trajectories needed to train offline RL policies from
observational registries without prospective randomization, attracting substantial
clinical interest following claims of 10--32\,\% improvements over physician
decisions~\citep{komorowski2018,choi2024,ghasemi2025}.


Several highly cited clinical RL papers share three methodological
limitations that can inflate apparent policy improvement:
(1)~reliance on direct Q-value evaluation rather than Fitted Q-Evaluation (FQE),
which we find overestimates policy value by 5--131$\times$ on our EHR data
(Figure~\ref{fig:confound}, inset), consistent with~\citet{tang2021};
(2)~no reward deconfounding, leaving prognostic severity signals embedded in the
proxy reward so that sicker patients who receive more intensive treatment drive
an apparent improvement signal; and
(3)~no E-value analysis~\citep{vanderweele2017}, providing no bound on the
strength of confounding required to explain the observed effect.

We characterize an RL-specific failure mode that falls outside the estimand
targeted by E-value sensitivity analysis.
Intuitively, the standard reward function based on 3-month modified Rankin Scale
(mRS) encodes how sick the patient was at baseline rather than how well the
treatment worked: sicker patients systematically receive more intensive treatment
and have worse outcomes, and the resulting value estimate can mistake this
severity--outcome correlation for a treatment effect.
Formally, we term this \emph{reward-embedded confounding}: the proxy reward
carries baseline severity/prognosis information in addition to any
treatment-efficacy signal, so the RL value estimate may mistake the
severity--outcome pathway for treatment benefit.
Crucially, \rend's estimate passes an \emph{external} unmeasured-confounding
sensitivity check (E-value $= 18.37$), yet a
$2{\times}2$ factorial experiment reveals that terminal reward confounding alone
exceeds the full signal magnitude --- i.e., removing terminal confounding
more than eliminates the observed improvement, because the raw 3-month mRS
encodes the severity$\to$outcome pathway that gradient boosting machine (GBM)
residualization removes (Figure~\ref{fig:factorial}).
The E-value is therefore necessary but not sufficient here: it targets
\emph{external} confounding, whereas this bias is carried through the reward
channel rather than only through treatment assignment.


This paper uses the offline RL evaluation pipeline as a diagnostic instrument
--- not to deploy a clinical policy, but to systematically expose evaluation
pitfalls that can produce premature positive policy-improvement claims.
We make five contributions:
\begin{enumerate}[leftmargin=*, nosep]
  \item \textbf{Reward-embedded confounding}: We formalize and decompose this
    failure mode via $2{\times}2$ factorial experiments, showing terminal
    confounding accounts for 218.6\,\% of the $\rend \to \rfull$ net change
    (Figure~\ref{fig:factorial}). A T-learner provides non-RL triangulation
    through substantial attenuation and a clinically small residual signal.

  \item \textbf{Limits of E-value analysis for policy improvement}: We apply
    E-value sensitivity analysis to policy improvement estimates, demonstrating
    that external-confounding sensitivity checks are
    insufficient to address reward-embedded confounding (the FQE signal falls from
    $+0.0101$ to $+0.0025$, $p=0.291$, under full deconfounding).

  \item \textbf{6-step evaluation checklist}: An empirically motivated checklist
    with pre-specifiable diagnostic checks~\citep{gottesman2019}
    (Table~\ref{tab:protocol}). Applied retrospectively, three highly cited
    clinical RL papers do not report the FQE confirmation specified by Step~1.

  \item \textbf{Largest-scale FQE vs.\ direct Q-evaluation}: Largest systematic
    quantification of direct Q overestimation (5--131$\times$ across
    ${\sim}$450 training runs), confirming~\citet{tang2021} on real EHR data.

  \item \textbf{From null to actionable}: Stratification by National Institutes
    of Health Stroke Scale (NIHSS) severity reveals a $4.6\times$ conditional
    average treatment effect (CATE) gradient for hypothesis generation and
    prospective randomized controlled trial (RCT) design.
\end{enumerate}


\subsection*{Related Work}
\label{sec:rw}

Deep RL for clinical decision-making gained prominence with
\citet{komorowski2018}, whose AI Clinician reported weighted importance
sampling (WIS)-estimated survival gains for intensive care unit (ICU) sepsis;
\citet{choi2024} and \citet{ghasemi2025} followed with
claims of 10-percentage-point and $+32\,\%$ improvements, respectively.
Methodological audits have since questioned whether these gains reflect true
treatment-benefit signals~\citep{luo2024}.
\citet{tang2021} demonstrated that FQE achieves Spearman $\rho = 0.89$ for
policy ranking while direct Q-value estimates achieve $\rho < 0.3$;
our experiments corroborate this on 44,894 real EHR trajectories
(5--131$\times$ overestimation).
Action imbalance further compounds the problem: \citet{nambiar2023kdd}
quantified ratios up to 6,400:1, rendering importance sampling (IS)-based
off-policy evaluation (OPE) unreliable before confounding is considered.

Prior work on confounding in sequential OPE~\citep{namkoong2020,kallus2020,lu2018deconfounding}
addresses only \emph{external} unmeasured confounders;
our focus is the reward-definition problem that arises when an observational
clinical outcome is used as the terminal reward.
Confounding by indication is well-established in
pharmacoepidemiology~\citep{salas1999},
and reward shaping theory establishes conditions under which reward
transformations preserve optimal policies~\citep{ng1999rewardshaping} ---
but when the reward itself encodes a confounding pathway,
these invariance guarantees do not apply.
This mechanism is also related to proxy-reward misspecification: a clinically
convenient proxy endpoint may optimize a prognostic correlate rather than the
treatment-relevant outcome of interest.
We term this RL-specific failure mode \emph{reward-embedded confounding} and
provide an empirical factorial decomposition.
Double/Debiased Machine Learning (DML)~\citep{chernozhukov2018} enables causal
estimation from observational data; we adapt its outcome-partialling logic to
prognosis-residualized reward evaluation as a diagnostic stress test, not as a
standalone causal treatment-benefit estimand (see \S\ref{sec:methods:reward}).

\citet{gottesman2019} identified seven evaluation pitfalls for RL in healthcare,
including inappropriate OPE selection and insufficient confounding analysis.
The three highly cited clinical RL publications we
review~\citep{komorowski2018,choi2024,ghasemi2025} --- including one predating
\citeauthor{gottesman2019}'s recommendations --- do not report the FQE
confirmation specified by Step~1 of our checklist.
Our 6-step checklist operationalizes \citeauthor{gottesman2019}'s recommendations
with concrete numerical diagnostics; external validation of this checklist is
future work.

\subsection*{Generalizable Insights about Machine Learning in the Context of Healthcare}

\begin{enumerate}[nosep, leftmargin=*]
  \item \textbf{Audit the reward channel separately.}
    When an observational outcome defines the reward, baseline prognosis can remain
    in the value after standard OPE. Raw-versus-residualized factorials
    test this pathway, which is distinct from treatment-assignment confounding;
    residualization is a diagnostic stress test, not a causal remedy.
  \item \textbf{Align endpoints with the treatment mechanism.}
    Functional outcome can be a weak proxy for secondary prevention; our direct
    AIPW analysis found no significant reduction in ischemic-stroke recurrence
    ($p=0.12$--$0.23$), without implying that every clinical endpoint was null.
  \item \textbf{Match model complexity to the data.}
    With 98.3\,\% state invariance, simpler causal methods may be more appropriate
    than a sequential policy model.
  \item \textbf{Pre-specify layered evaluation checks.}
    Our six-step checklist combines FQE confirmation, support, external- and
    reward-channel confounding, non-RL triangulation, and subgroup stability.
    It is motivated by one Korean registry; external validation remains future work.
\end{enumerate}

%% file: methods.tex

\section{Methods}
\label{sec:methods}


\subsection{Dataset}
\label{sec:methods:data}

We used the Clinical Research Collaboration for Stroke in Korea (CRCS-K) registry,
a prospective multicenter registry of consecutive ischemic stroke patients admitted
to 20 tertiary hospitals across South Korea from 2008 to 2023~\citep{crcs-k}.
Registry data use was approved under the institutional research agreement governing
CRCS-K multicenter data sharing. Registry enrollment and research use of clinical
data were approved by the institutional review boards of the participating centers,
with written informed consent obtained from patients or their legally authorized
representatives. The present secondary analysis of deidentified CRCS-K registry
data was approved by the Institutional Review Board of Seoul National University
Bundang Hospital (IRB No.~\mbox{B-2308-845-302}).
For main analyses we restricted to post-2018 admissions ($N = 44{,}894$),
following the 2018 Korean Stroke Society guideline update that standardized
antithrombotic practice (Appendix~\ref{app:temporal}).
Functional outcome was measured by 3-month modified Rankin Scale (mRS, 0--6;
90.2\,\% coverage; the analyzed cohort has 100\,\% by construction).
Three cohort sizes appear throughout: \textbf{$N = 44{,}894$} (primary RL cohort);
\textbf{$N = 44{,}974$} (T-learner cohort, 80 additional patients with valid mRS
but incomplete MDP episodes; Appendix~\ref{app:tlearner});
\textbf{$N = 35{,}744$} (1-year mRS subcohort;
Appendix~\ref{app:mrs1y_coverage}).
Cohort demographics are in Appendix Table~\ref{tab:cohort}.

We defined six antithrombotic treatment actions from discharge medication records
(distributions in Appendix Table~\ref{tab:cohort}):
dual antiplatelet (AP\_dual, 58.9\,\%), anticoagulant+statin (AC\_statin, 16.2\,\%),
antiplatelet+statin (AP\_statin, 15.1\,\%), conservative (5.6\,\%), anticoagulation
alone (AC, 2.5\,\%), and antiplatelet monotherapy (AP\_mono, 1.7\,\%).
Statin co-prescription is included because it co-varies systematically with
antithrombotic choice in post-stroke prescribing (statin rates differ by
$>$20 percentage points across antithrombotic categories).
This 35:1 action imbalance has direct implications for IS validity
(\S\ref{sec:methods:eval}; Appendix~\ref{app:positivity}), consistent with~\citet{nambiar2023kdd}.


\subsection{MDP Formulation}
\label{sec:methods:mdp}

We formulated a 2-step episodic Markov Decision Process (MDP)
$\mathcal{M} = (\mathcal{S}, \mathcal{A}, \mathcal{T}, \mathcal{R}, \gamma)$
corresponding to the acute ($t=0$) and discharge ($t=1$) phases.
The \emph{state space} $\mathcal{S} \subset \mathbb{R}^{421}$ consists of
421 clinical features (admission demographics, 14 lab values, NIHSS sub-items,
pre-stroke medication history, AI imaging features;
Appendix~\ref{app:features}).
The \emph{action space} $\mathcal{A} = \{0, \ldots, 5\}$ (hereafter Config~A)
is the six antithrombotic strategies defined above.
The discount factor is $\gamma = 0.99$.
The \emph{reward function} $\mathcal{R}$ defines 14 variants
(\S\ref{sec:methods:reward}), six of which are central to our analysis.
An anchor experiment confirms that the correction from an earlier 3-step to
this 2-step formulation does not materially affect results
($\vimp = +0.0024$ vs.\ $+0.0020$, $p = 0.94$;
Appendix~\ref{app:mdp_anchor}).


\subsection{Offline RL Algorithms}
\label{sec:methods:algo}

\begin{sloppypar}
We trained five offline RL algorithms: Conservative Q-Learning
(CQL,~$\alpha \in \{0.5,1,2,4,8\}$)~\citep{kumar2020cql} as the primary algorithm,
Batch-Constrained Q-Learning (BCQ,~$f \in \{0.1,0.3,0.5\}$)~\citep{fujimoto2019bcq},
Implicit Q-Learning (IQL,~$\tau \in \{0.7,0.8,0.9\}$)~\citep{kostrikov2022iql},
Decision Transformer (DT)~\citep{chen2021dt}, and Behavior Cloning (BC).
\end{sloppypar}
In the common-reward screen, none of the five families showed positive
improvement evidence (Table~\ref{tab:main_results}; Appendix~\ref{app:heatmap}):
CQL, BCQ, and BC FQE intervals included zero, IQL FQE was below physicians,
and DT's direct value estimate collapsed to the physician-value baseline
(FQE was unavailable for DT). The design was partially crossed:
post-2018 reward deconfounding used only the prespecified primary CQL
$\alpha=4.0$, selected for FQE TD-error convergence across reward variants.
Training used \texttt{d3rlpy} v2.0~\citep{seno2021d3rlpy} on one NVIDIA H100
(${\approx}350$ GPU-hours; hyperparameters in Appendix~\ref{app:hyperparams});
all 33 experiments are mapped in Appendix~\ref{app:evidence_map}.


\subsection{Reward Design}
\label{sec:methods:reward}

We designed 14 reward variants following a progressive confounding-removal
strategy.
Eight preliminary variants (raw mRS, NIHSS-adjusted, binary, severity-stratified,
etc.) are superseded by the six variants central to our analysis:
\begin{itemize}
  \item \textbf{\rstd} (standard evaluation): intermediate reward
    $r_t = f(\text{NIHSS}_{24h}, \text{NIHSS}_0)$ (missing in ${\approx}80\,\%$
    of records; defaults to zero when absent) and terminal reward
    $r_T = -\text{mRS}_{3\text{mo}}/6 - 0.3 \cdot \mathbb{1}[\text{death}]
    + 0.1 \cdot \mathbb{1}[\text{mRS} \leq 1]$.
    This represents typical practice in clinical RL.

  \item \textbf{\rdml} (DML reward residualization): terminal reward replaced by
    $r_T = -(\text{mRS}_{3\text{mo}} - \hat{f}(\mathbf{x}_0)) / 6$, where
    $\hat{f}$ is a LightGBM model trained on 743 treatment-free baseline features
    via cross-fitting ($R^2 = 0.70$, calibration slope $= 1.03$,
    zero treatment variables in SHAP top-20).
    We adopt the term \rdml{} for conciseness, noting that our approach
    applies only the outcome-residualization component of Double/Debiased
    ML~\citep{chernozhukov2018}; treatment residualization is omitted
    because near-universal statin prescription renders the treatment-partialling
    step numerically degenerate. We use the residualized reward as a
    prognosis-residualized diagnostic stress test, not as a standalone causal
    treatment-benefit estimand (details in Appendix~\ref{app:causal_structure}).
    Semi-synthetic calibration suggests 24--26\,\% absorption, while a more
    conservative propensity-based bound gives 27\,\%; main-text claims use the
    conservative bound (Appendix~\ref{app:attenuation}).

  \item \textbf{\rend} ($\lambda = 0.2$): \rstd{} terminal reward augmented with
    intermediate penalty for early neurological deterioration (END):
    $r_0 = r_0^{\text{std}} - \lambda \cdot \text{END}_{\text{flag}}$,
    where END is $\geq$4-point NIHSS worsening within 72 hours.

  \item \textbf{\rfull} (full deconfounding): residualizes both reward
    components against baseline prognosis. The terminal reward is identical to
    \rdml, $r_T = -(\text{mRS}_{3\text{mo}} - \hat{f}(\mathbf{x}_0))/6$.
    The intermediate reward is
    $r_0 = r_{0,\text{std}} - \lambda(\text{END}_{\text{flag}} -
    \hat{g}(\mathbf{x}_0))$, where $\hat{g}$ predicts END from the same
    treatment-free baseline features. For the factorial decomposition,
    \rendprime{} denotes the cell with the \rend{} intermediate reward and
    the \rdml{} terminal reward.

  \item \textbf{\ryr} (1-year outcome): terminal reward
    $r_T = -(\text{mRS}_{1\text{yr}} - \bar{m}_b) / 6$,
    where $\bar{m}_b$ is the cohort-mean mRS$_{1\text{yr}}$ used to center
    the reward, with NIHSS 24-hour intermediate reward.
    Uses 35{,}744 patients with 1-year follow-up.

  \item \textbf{\ryrdml} (DML-residualized 1-year outcome): terminal reward
    $r_T = -(\text{mRS}_{1\text{yr}} - \hat{g}(\mathbf{x}_0)) / 6$,
    where $\hat{g}$ is a LightGBM model predicting $\text{mRS}_{1\text{yr}}$
    (cross-fit $R^2 = 0.75$, calibration slope $= 1.03$,
    zero treatment variables in SHAP top-20).
    \ryr{} and \ryrdml{} form a second factorial pair crossing outcome horizon
    with deconfounding.
\end{itemize}

The diagnostic mechanism is that baseline severity $\mathbf{X}$ causes both
treatment $T$ and outcome $Y$; the standard reward $r = f(Y)$ conflates the
prognostic ($\mathbf{X} \to Y$) and treatment-attributable ($T \to Y$) pathways.
GBM residualization subtracts baseline prognosis and tests whether the apparent
policy advantage survives removal of this pathway. SHAP validation confirms zero
treatment variables in the GBM top-20 (Appendix~\ref{app:causal_structure}).


\subsection{Evaluation Hierarchy}
\label{sec:methods:eval}

We adopt a six-step evaluation checklist (Table~\ref{tab:protocol}).
As the primary off-policy evaluation method we use \textbf{Fitted Q-Evaluation
(FQE)}~\citep{tang2021}, which fits a separate Q-function to evaluate a fixed
target policy.
All results use 15-fold cross-validation (5 folds $\times$ 3 random seeds).
The primary metric is $\vimp = \vrl - \vphy$ evaluated under FQE.

To quantify confounding robustness we apply the E-value
framework~\citep{vanderweele2017} to the observed $\vimp$.

\textbf{Structural limitation.}
The E-value addresses only \emph{external} unmeasured confounders ($U \to T$, $U \to Y$).
Reward-embedded confounding operates through observed baseline severity
\emph{encoded within the reward itself} --- a reward-channel pathway that the
E-value does not evaluate.
The $2{\times}2$ factorial decomposition (\S\ref{sec:results:act2}) serves as
the primary diagnostic.

\begin{sloppypar}
\textbf{Approximate computation.}
We map $\vimp$ to RR scale via
$\widehat{RR} = \exp(|\vimp| / \text{SE}_{\text{cons}})$, where
$\text{SE}_{\text{cons}} = \text{SD}_{\text{folds}} / \sqrt{3}$ (conservative).
This transformation provides a conservative upper bound on confounding
strength required to explain away $\vimp$; derivation and justification
appear in Appendix~\ref{app:evalue_details}.
\end{sloppypar}

We employ four additional triangulation strategies (doubly robust OPE
(DR-OPE), instrumental variable (IV), difference-in-differences (DiD),
T-learner / direct recurrence) detailed in
\S\ref{sec:results:act2} and
Appendices~\ref{app:positivity}--\ref{app:recurrence}.


\subsection{Cross-validation and Statistical Testing}
\label{sec:methods:cv}

All experiments use patient-level 5-fold cross-validation repeated across 3
random seeds (15 evaluations per configuration;
see Appendix~\ref{app:cv_details} for cross-validation details).
Significance of $\vimp$ is assessed via one-sample $t$-test against zero
($\text{df} = 14$); factorial effects via paired $t$-tests ($N = 15$ pairs).
We report 95\,\% CI with $\alpha = 0.05$; Bonferroni correction across
14 reward variants ($\alpha_{\text{adj}} = 0.0036$) would leave \rend{}
significant but not \rstd{} --- our argument does not depend on \rstd{}
significance.
Power analysis and the Nadeau--Bengio variance correction are discussed in
\S\ref{sec:discussion:power}.

%% file: results.tex

\section{Results}
\label{sec:results}

We present results in three acts that mirror the diagnostic sequence,
followed by an analysis of clinical heterogeneity.
All $\vimp = \vrl - \vphy$ estimates are based on FQE with
15-fold CV unless stated otherwise.
Table~\ref{tab:main_results} compactly summarizes the algorithm screen and six
key reward variants.

\begin{table}[t]
  \caption{Compact evaluation robustness summary: a common-reward screen across
           five algorithm families, followed by six FQE reward designs for primary CQL
           $\alpha{=}4.0$ (15-fold CV, 95\,\% CI from $t$-distribution, 14~df).
           E-values are approximate upper bounds computed via a non-standard
           continuous-to-RR mapping (\S\ref{sec:methods:eval});
           Bonferroni-corrected $\alpha_{\text{adj}} = 0.0036$ would leave
           \rend{} significant but not \rstd{}.}
  \label{tab:main_results}
  \centering
  \small
  \begin{tabular}{lp{4.2cm}cccc}
    \toprule
    Reward & Description &
    $\vimp$ & 95\,\% CI & $p$ & E-value \\
    \midrule
    Five families & Common reward (FQE except DT; Appendix~\ref{app:heatmap}) &
    \multicolumn{4}{c}{No positive evaluated evidence} \\
    \midrule
    \multicolumn{6}{l}{\textit{3-month outcome ($N{=}44{,}894$)}} \\
    \rstd & NIHSS-bin FQE (standard) &
    $+0.0069$ & $[+0.0001,\ +0.014]$ & $0.048$  & 4.70 \\
    \rend & +END intermediate ($\lambda{=}0.2$) &
    $+0.0101$ & $[+0.006,\ +0.014]$ & $0.0002$ & 18.37 \\
    \rdml & GBM residualization (DML) &
    $+0.0033$ & $[-0.001,\ +0.008]$ & $0.132$  & 3.50 \\
    \rfull & Full deconfound.\ (\rdml{} + END resid.) &
    $+0.0025$ & $[-0.002,\ +0.007]$ & $0.291$  & 2.65 \\
    \midrule
    \multicolumn{6}{l}{\textit{1-year outcome ($N{=}35{,}744$)}} \\
    \ryr & NIHSS-bin, mRS$_{1\text{yr}}$ terminal &
    $+0.0133$ & $[+0.006,\ +0.020]$ & $0.001$  & 11.44 \\
    \ryrdml\textsuperscript{\dag} & GBM residualized mRS$_{1\text{yr}}$ &
    $-0.0004$ & $[-0.005,\ +0.004]$ & $0.843$  & ---\textsuperscript{\ddag} \\
    \bottomrule
    \multicolumn{6}{@{}p{\textwidth}@{}}{\footnotesize
      \textsuperscript{\dag}\ryrdml{} results are from Phase~12 (post-2018 cohort restricted to
      2018--2023 admissions with 1-year follow-up, $N = 35{,}744$).
      \textsuperscript{\ddag}E-value not computed ($\vimp < 0$; E-value is defined only for
      positive point estimates).} \\
  \end{tabular}
\end{table}


\subsection{Act 1: Standard Offline RL Evaluation Yields Positive Signal}
\label{sec:results:act1}

MDP structure diagnosis leaves FQE as the least problematic primary evaluator
here, not an assumption-free one (including Q-function realizability;
\S\ref{sec:discussion:limitations}): IS-ESS $= 0.21\,\%$ ($48\times$ below
threshold) rules out IS-based OPE
(Appendix~\ref{app:mdp_diagnosis}; direct Q overestimates FQE by
5--131$\times$ across ${\sim}$450 runs; Figure~\ref{fig:confound}, inset).

Under standard practice, \rstd{} produces $\vimp = +0.0069$
(95\,\%\,CI $[+0.0001, +0.014]$, $p = 0.048$, E-value $= 4.70$).
Adding an END intermediate reward (\rend, $\lambda = 0.2$) strengthens the
signal to $\vimp = +0.0101$ ($p = 0.0002$, E-value $= 18.37$) --- an estimate
that a reader applying standard practice could interpret as strong evidence for
policy improvement.

Direct Q-evaluation yielded estimates 5--131$\times$ larger than FQE across
${\sim}$450 runs (Figure~\ref{fig:confound}, inset),
consistent with~\citet{tang2021}.
As a rough illustration, if our overestimation factor generalized,
studies reporting $\vimp = +0.10$ to $+0.32$ via direct Q would report
$+0.001$ to $+0.003$ under FQE --- though this extrapolation is speculative,
as the ratio is dataset- and algorithm-dependent.
The natural question is whether the positive FQE signal reflects treatment
benefit or reward-embedded confounding; Act~2 examines this directly.


\subsection{Act 2: Progressive Deconfounding Attenuates the Signal to Null}
\label{sec:results:act2}

GBM reward residualization (\rdml) replaces the raw terminal reward with
the residual after subtracting LightGBM-predicted prognosis
(cross-fit $R^2 = 0.70$, calibration slope $= 1.03$,
0 treatment variables in SHAP top-20).
The signal attenuates: $\vimp = +0.0033$
(95\,\%\,CI $[-0.001, +0.008]$, $p = 0.132$), a 52\,\% reduction from \rstd.
Over-adjustment risk is bounded conservatively at 27\,\%
(Appendix~\ref{app:attenuation}), and the attenuation-adjusted estimate remains
clinically small ($\vimp_{\mathrm{adj}}\approx +0.0045$, $\leq 2.7\,\%$ MCID).

\begin{figure}[t]
  \centering
  \includegraphics[width=\textwidth]{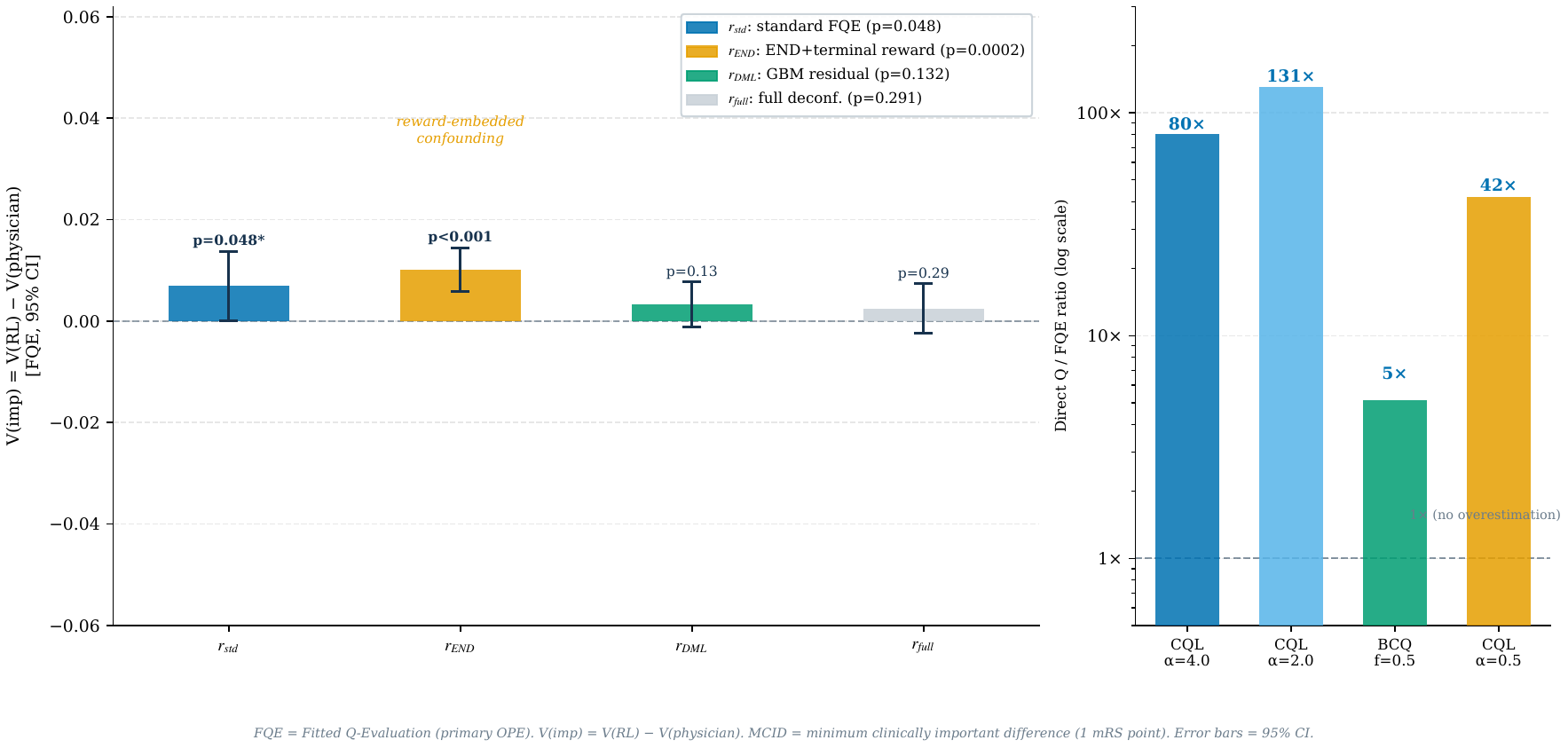}
  \caption{%
    \textbf{Confounding cascade across reward designs.}
    FQE-based $\vimp$ for four reward variants (CQL $\alpha{=}4.0$, 15-fold CV).
    The apparent improvement attenuates progressively as prognostic confounding
    is removed ($\rstd \to \rdml \to \rfull$).
    \rend{} ($+0.0101$, $p = 0.0002$, E-value $= 18.37$) falls to
    $\rfull{}=+0.0025$ ($p=0.291$) under full reward deconfounding
    (\S\ref{sec:results:act2}).
    \textit{Inset:} Direct Q overestimates FQE by 5--131$\times$ (log scale).
    Error bars: 95\,\% CI.
  }
  \label{fig:confound}
\end{figure}

Full deconfounding (\rfull) yields $\vimp = +0.0025$
(95\,\%\,CI $[-0.002, +0.007]$, $p = 0.291$), consistent with a null effect.
Figure~\ref{fig:confound} shows the complete confounding cascade:
$\rstd$ ($+0.007$) $\to$ $\rdml$ ($+0.003$) $\to$ $\rfull$ ($+0.002$).

\begin{figure}[t]
  \centering
  \includegraphics[width=\textwidth]{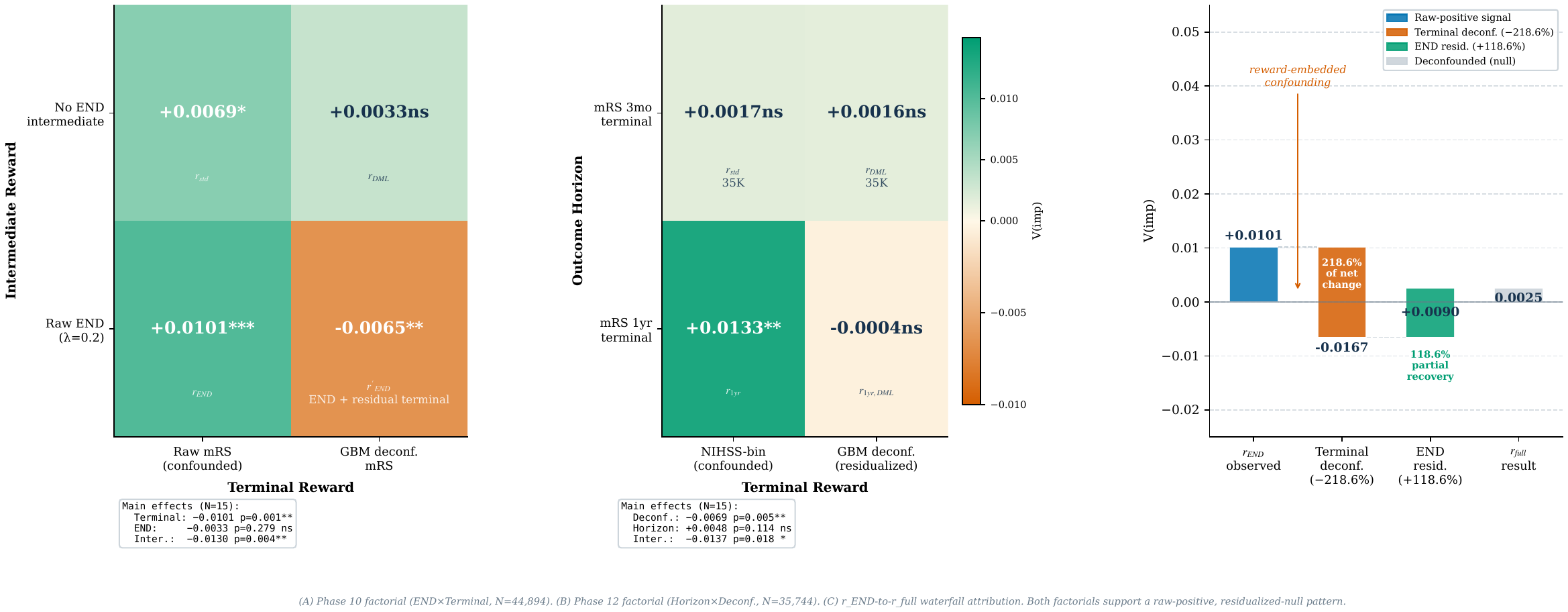}
  \caption{%
    \textbf{Dual $2{\times}2$ factorial decomposition.}
    \textit{(A) Phase 10:} Crossing terminal reward (raw vs.\ GBM-residualized mRS$_{3\text{mo}}$)
    and END intermediate handling on $N{=}44{,}894$.
    Terminal deconfounding main effect: $-0.0101$ ($p = 0.001$).
    \textit{(B) Phase 12:} Crossing outcome horizon with deconfounding on $N{=}35{,}744$.
    Deconfounding main effect: $-0.0069$ ($p = 0.005$); \ryr's $+0.0133$ reduces
    to $-0.0004$ after GBM residualization (103.3\,\% attenuation).
    \textit{(C) Waterfall:} Terminal confounding accounts for \textbf{218.6\,\%}
    of the net $\rend \to \rfull$ signal change
    ($\rend{=}+0.0101$, $\rendprime{=}-0.0065$, $\rfull{=}+0.0025$).
  }
  \label{fig:factorial}
\end{figure}

To identify the mechanism driving \rend's apparent $+0.010$ signal,
we designed a $2{\times}2$ factorial experiment crossing terminal reward
(raw mRS vs.\ GBM-residualized) and intermediate reward
(raw END vs.\ GBM-residualized).
Here $\rendprime$ denotes the factorial cell that combines \rend's
END-augmented intermediate reward with a GBM-residualized terminal reward.
The decomposition reveals the magnitude of reward-embedded confounding:
the simple effect of terminal residualization within the END-augmented reward
($\rend \to \rendprime$) is $\Delta\vimp = -0.0167$,
\textbf{accounting for 218.6\,\% of the net $\rend \to \rfull$ change}
(computed from unrounded fold-level estimates:
$\frac{\vimp(\rendprime) - \vimp(\rend)}{\vimp(\rfull) - \vimp(\rend)}
= \frac{-0.01668}{-0.00763} = 218.6\,\%$).
The percentage exceeds 100\,\% because terminal residualization alone overshoots
null; the residualized intermediate reward partially offsets this
($\Delta\vimp = +0.0090$, $p = 0.279$).
The main effect of terminal deconfounding is $-0.0101$ ($p = 0.001$);
the significant interaction ($-0.0130$, $p = 0.004$) indicates that
END amplifies terminal confounding through a shared prognostic pathway
(severity $\to$ END $\to$ mRS; Figure~\ref{fig:factorial}).

A second factorial (Phase~12, $N = 35{,}744$) crosses outcome horizon
(mRS$_{3\text{mo}}$ vs.\ mRS$_{1\text{yr}}$) with deconfounding.
The 1-year reward (\ryr) produces $\vimp = +0.0133$ ($p = 0.001$,
E-value $= 11.44$); GBM residualization (\ryrdml) attenuates the signal by
\textbf{103.3\,\%}, reducing to $-0.0004$ ($p = 0.843$).
The deconfounding main effect ($-0.0069$, $p = 0.005$; from fold-level
unrounded estimates) supports the interpretation that the raw-positive,
residualized-null pattern is not specific to the
3-month mRS horizon in this registry (Figure~\ref{fig:factorial}B).

\rend's E-value of 18.37 is mathematically correct for \emph{external}
unmeasured confounders within that sensitivity framework.
The confounding is instead \emph{internal} to the reward function:
the raw 3-month mRS encodes the severity$\to$outcome prognostic pathway
that GBM residualization removes.
We term this \emph{reward-embedded confounding}: the proxy reward carries a
non-treatment-attributable prognostic pathway, so the learned policy can appear
favorable even when the FQE signal is not robust to reward deconfounding.
E-values remain informative for external unmeasured confounding but do not
address this reward-channel pathway; factorial decomposition or
prognosis-residualized reward evaluation is needed as a diagnostic stress test.

\paragraph{Causal triangulation.}
\label{sec:results:tlearner}%
\label{sec:results:recurrence}%
Two independent approaches corroborate the deconfounding result.
A T-learner~\citep{kunzel2019} ($N = 44{,}974$;
Appendix~\ref{app:tlearner}) shows 75\,\% signal attenuation under
\rdml{} residualization in the full cohort and 90\,\% in the overlap zone
($0.1 < e(X) < 0.9$, $n=9{,}900$), with a clinically negligible residual
($< 6\,\%$ MCID).
Direct recurrence analysis via inverse probability weighting (IPW) /
augmented IPW (AIPW) ($N = 32{,}583$) finds no significant
ischemic stroke recurrence reduction for any treatment comparison
(AIPW $p=0.12$--$0.23$; Appendix~\ref{app:recurrence}).
DR-OPE serves as a consistency check; IV is uninformative
(exclusion restriction violation; Appendix~\ref{app:iv});
DiD excluded for parallel trends violation (Appendix~\ref{app:did}).


\subsection{MDP Structure Diagnosis}
\label{sec:results:mdp}

MDP structure analysis confirms that 98.3\,\% of state features are invariant
across timesteps (Pearson $r > 0.99$), with per-patient Q-value rank
$\rho = 0.82$ (Figure~\ref{fig:statecollapse}, left). IS-based OPE is infeasible
(ESS $= 0.21\,\%$, $48\times$ below threshold; Figure~\ref{fig:statecollapse},
right), leaving FQE as the least problematic OPE method in this setting
(see Appendix~\ref{app:mdp_diagnosis}).


\subsection{Clinical Heterogeneity: Hypothesis Generation for Prospective Evaluation}
\label{sec:results:hetero}

Having established that the overall policy improvement is null after reward
deconfounding, we turn to whether clinically relevant heterogeneity remains as a
trial-design signal --- not to claim treatment benefit, but to identify subgroups
for prospective evaluation.

\begin{figure}[t]
  \centering
  \includegraphics[width=\textwidth]{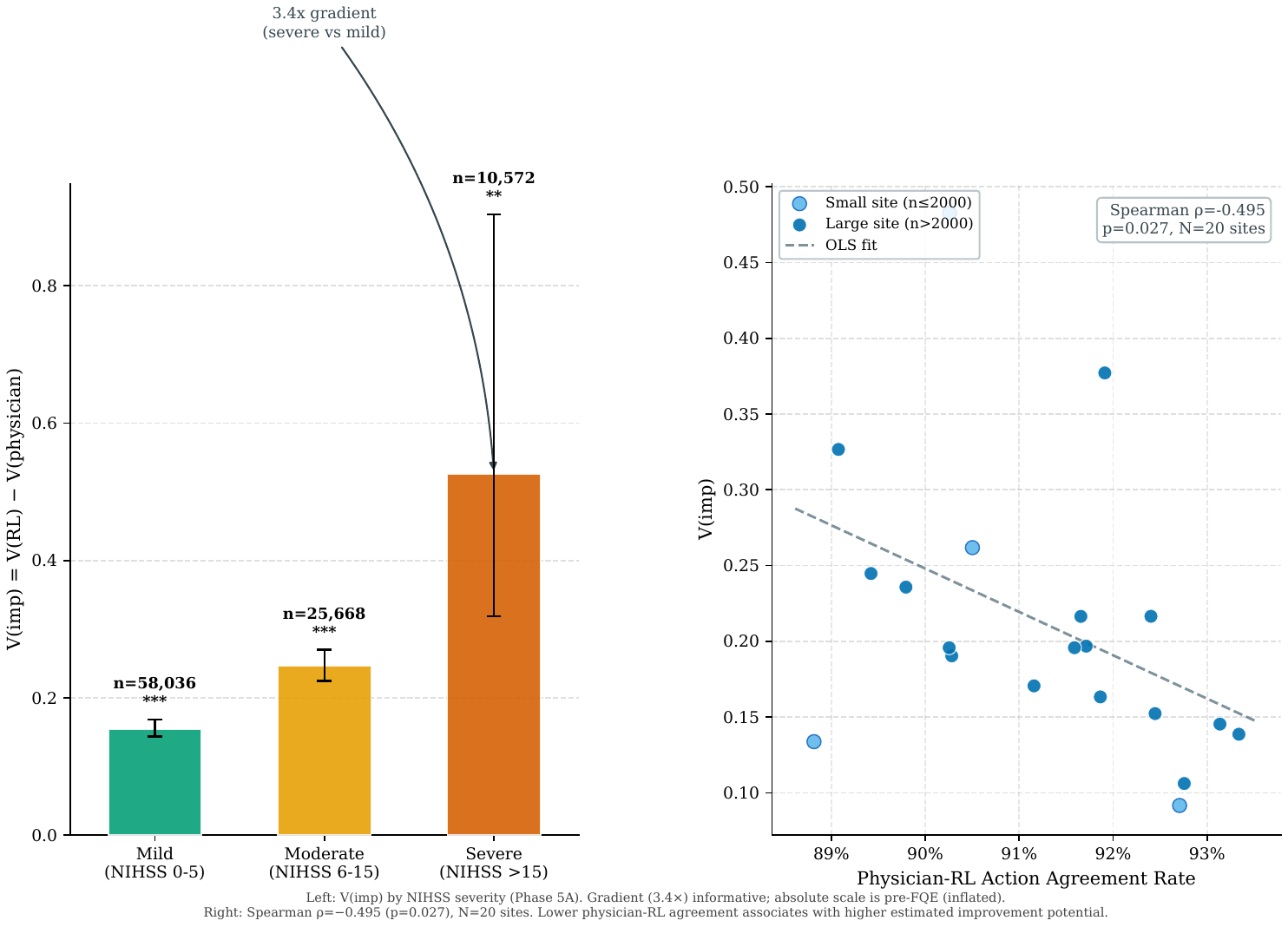}
  \caption{%
    \textbf{Hypothesis-generating clinical heterogeneity after deconfounding diagnostics.}
    \textit{Left:} $\vimp$ by NIHSS severity (Phase~5A direct Q-evaluation,
    $N = 116{,}215$; absolute values unreliable but $3.4\times$ relative gradient
    is directionally consistent with the T-learner's $4.6\times$ CATE ratio).
    \textit{Right:} Hospital-level $\vimp$ vs.\ physician--RL agreement
    across 20 tertiary centers.
    The raw association is hypothesis-generating only and does not persist under
    \rfull{} (see \S\ref{sec:results:hetero}).
  }
  \label{fig:heterogeneity}
\end{figure}

Despite the overall null result after deconfounding, the NIHSS severity gradient
persists as an exploratory, hypothesis-generating pattern.

The T-learner identifies a \textbf{4.6$\times$ severe/mild CATE ratio}
across NIHSS severity subgroups, directionally consistent with a $3.4\times$
$|\vimp|$ gradient under direct Q-evaluation (Figure~\ref{fig:heterogeneity}).
Directional consistency across two independent methods measuring different
estimands (CATE vs.\ policy value) supports hypothesis-generating clinical
heterogeneity rather than relying on a single policy-value estimate.
Hospital-level $\vimp$ under direct Q-evaluation correlates negatively with
physician--RL agreement ($\rho = -0.495$, $p = 0.027$, $N = 20$;
jackknife direction-stable; Figure~\ref{fig:heterogeneity}); however,
this correlation vanishes under the deconfounded reward
($\rho \approx 0$, ns; recomputed estimate $\rho = -0.026$, $p = 0.915$),
suggesting an unstable, confounding-driven site signal rather than surviving
site-level practice variation.
We therefore focus heterogeneity interpretation on the NIHSS-stratified
T-learner estimates as the more reliable estimand.
Policy disagreement profiling reveals that 57\,\% of RL--physician
disagreements involve statin co-prescription, with AF history as the strongest
driver (Appendix~\ref{app:disagreement}).
These patterns suggest that antithrombotic--statin combination decisions in
severe-to-moderate NIHSS patients constitute the highest-priority subgroup
for prospective evaluation.

%% file: protocol.tex

\section{A 6-Step Evaluation Checklist for Clinical Offline RL}
\label{sec:protocol}

Drawing on the diagnostic sequence revealed in our study, we propose a
six-step empirical evaluation checklist that operationalizes~\citet{gottesman2019}'s
seven evaluation pitfalls with pre-specifiable diagnostic checks
(Table~\ref{tab:protocol}). The checklist is intended to identify when a
positive $\vimp$ estimate needs additional evidence before it is interpreted
as clinical benefit; external validation of the checklist remains future work.

\begin{table}[t]
  \caption{6-Step Evaluation Checklist for Clinical Offline RL.
           \cmark~=~Met; \xmark~=~Not met or not assessable;
           NR~=~Not reported; NI~=~Not implemented.
           Steps 1--2: OPE reliability;
           Steps 3--4: confounding robustness;
           Steps 5--6: causal triangulation and generalizability.}
  \label{tab:protocol}
  \centering
  \small
  \setlength{\tabcolsep}{2pt}
  \resizebox{\textwidth}{!}{%
  \begin{tabular}{@{}clp{2.8cm}p{3.8cm}p{3.4cm}@{}}
    \toprule
    Step & Check & Threshold & This Study & Existing Papers \\
    \midrule
    S1 & FQE validation &
      Report FQE; if $|V_\text{Dir} / V_\text{FQE}| \geq 2\times$, do not interpret direct Q alone &
      \cmark~Ratio $= 5$--$131\times$ $\Rightarrow$ FQE primary &
      NR: WIS only~\citep{komorowski2018,ghasemi2025}; matching~\citep{choi2024}; no FQE \\
    \addlinespace
    S2 & IS viability (ESS $\geq$ 10\,\%) &
      ESS ratio $\geq 0.10$ &
      \xmark~\rstd{} ESS${=}0.21\%$ ($48{\times}$ below threshold); \rdml{} ESS${=}2.5\%$\textsuperscript{b} $\Rightarrow$ FQE least problematic &
      NR \\
    \addlinespace
    S3 & Confounding robustness (E-value) &
      Report E-value; necessary but not sufficient --- see S4 &
      \cmark~\rend{} $E{=}18.37$ (passes external-confounding check but falls under S4; \S\ref{sec:results:act2}); \xmark~\rstd{} $E{=}4.70$ (marginal) &
      NR \\
    \addlinespace
    S4 & Reward deconfounding &
      DML/GBM residual; $R^2 \geq 0.70$; 0 treatment vars in SHAP top-20 &
      \cmark~$R^2{=}0.70$, calibration $= 1.03$, 0 treatment vars &
      NI \\
    \addlinespace
    S5 & Causal triangulation &
      $\geq 2$ independent methods agree on direction &
      \cmark~FQE $+$ GBM null; T-learner attenuation 75\,\% full cohort,
      90\,\% overlap-trimmed ($n{=}9{,}900$); IS recurrence AIPW ns
      ($p{=}0.12$--$0.23$) &
      NI \\
    \addlinespace
    S6 & Subgroup stability &
      Pre-specified or hypothesis-generating subgroups; assess deconfounding stability &
      $\triangle$\textsuperscript{a}~NIHSS gradient $4.6\times$ (T-learner),
      $3.4\times$ (exploratory Q); site signal does not persist under \rfull{} &
      NI \\
    \bottomrule
  \end{tabular}}%
  \smallskip\\
  {\footnotesize NR = Not reported; NI = Not implemented.
   \textsuperscript{a}S6 CONDITIONAL: subgroups are exploratory;
   NIHSS gradient is a trial-design signal, not a treatment recommendation;
   site $\rho$ does not survive full reward deconfounding ($\rho \approx 0$,
   recomputed $\rho = -0.026$, $p = 0.915$ under \rfull).
   \textsuperscript{b}\rdml{} ESS$=2.5\%$ is the IS effective sample size ratio
   under the GBM-residualized reward policy, distinct from the AC action's
   2.5\% prescription rate (\S\ref{sec:methods:data}); both are coincidentally equal.}
\end{table}

Retrospective application to three widely cited clinical RL
papers~\citep{komorowski2018,choi2024,ghasemi2025} shows that Step~1 could
not be assessed as satisfied under this checklist: none reports FQE or
equivalent model-based OPE. Steps 4--6 were not reported in these papers,
leaving positive findings unexamined for reward-channel confounding.
We acknowledge this checklist was derived from diagnostic experience rather than
pre-registered or externally validated; external validation is future work.
The data and code availability statement specifies the permitted release
mechanism for analysis code, configuration files, figure-generation scripts, and
a synthetic example dataset designed to exercise the analysis pipeline without
exposing restricted patient-level registry records.

%% file: discussion.tex

\section{Discussion}
\label{sec:discussion}


\subsection{When Can Offline RL Add Value in Clinical Settings?}
\label{sec:discussion:when}

Our analysis suggests three conditions under which offline RL is most likely
to add value beyond simpler causal inference:
(a)~\textbf{reward-channel diagnostics} (proxy rewards should be
stress-tested for prognostic confounding; this is the diagnostic condition
emphasized here),
(b)~\textbf{meaningful sequential dynamics} (state features must evolve between
decision steps; our 98.3\,\% invariance limits optimization to near-1-step),
and (c)~\textbf{action support balance} (IS-ESS $\geq$ 10\,\% threshold;
our 2.5\,\% AC prescription rate drives IS-ESS collapse to
0.21\,\%, making FQE the least problematic OPE choice here).
In settings that satisfy all three, RL may provide value beyond simpler methods.
In our case, several key findings were discoverable \emph{only} through the RL
pipeline: the 5--131$\times$ FQE vs.\ direct Q gap, IS-ESS collapse, the factorial
decomposition, and the 6-step evaluation checklist.
Condition~(a) reflects what we term \emph{reward-embedded confounding}
(\S\ref{sec:results:act2}): a mechanism analogous to
confounding by indication in pharmacoepidemiology, where sicker patients
receive more aggressive treatment and have worse outcomes.
In classical OLS settings this makes treatment appear \emph{harmful};
in our offline RL setting the same severity$\to$outcome pathway is
encoded within the terminal reward, making the learned policy appear
\emph{favorable} --- the sign is reversed because the reward is
severity-correlated, not the propensity score.
In addition to confounding by indication in treatment assignment, the reward
itself carries the severity-to-outcome pathway, which is why this pathway falls
outside the E-value's external-confounding estimand.
For single-decision settings satisfying only condition~(a),
Causal Forest~\citep{wager2018} or DML~\citep{chernozhukov2018} directly
estimates CATE without sequential assumptions.


\subsection{Power Analysis and Effect Size Bounds}
\label{sec:discussion:power}

Our study has sufficient statistical power to detect clinically meaningful effects.
The retrospective minimum detectable effect (MDE) at 80\,\% power is
$\vimp \approx 0.010$ for \rstd{} and $\approx 0.006$ for \rdml,
corresponding to \textbf{5.9\,\%} and \textbf{3.8\,\%} of the mRS minimal
clinically important difference (MCID $= 1.0$ mRS point $= 0.167$ in $\vimp$
scale;~\citet{cranston2017}).
The Nadeau--Bengio correction for training-set overlap raises MDE to
$6.6\,\%$ MCID --- still below clinically meaningful thresholds.
Even taken at face value, the deconfounded estimates (\rdml{} $= +0.0033$,
\rfull{} $= +0.0025$) correspond to ${\approx}2.0$\,\% and $1.5$\,\% MCID
respectively --- well below the clinically relevant threshold we are powered
to detect ($> 3.8$\,\% MCID).
The null conclusion therefore reflects a statistically null and clinically
small estimated policy advantage, not insufficient power: we had sufficient
sensitivity to detect effects at or above this clinically meaningful range,
although smaller effects remain possible.

The near-null result reflects a specific property of well-managed tertiary
registries: in the post-2018 cohort, $>$80\,\% of AF patients receive
anticoagulation, reflecting near-universal guideline concordance that leaves
limited scope for RL to improve upon current practice.
This conclusion is specific to antithrombotic selection in Korean tertiary
stroke registries and does not imply that offline RL cannot improve stroke care
in settings with greater treatment heterogeneity, different treatment dimensions
(thrombolysis timing, endovascular intervention), or richer data modalities.


\subsection{Limitations}
\label{sec:discussion:limitations}

\begin{enumerate}[leftmargin=*, nosep]
  \item \textbf{External validity.} CRCS-K comprises predominantly Korean patients
    from tertiary referral centers with high guideline adherence;
    generalizability to other ethnic populations, community hospitals,
    and healthcare systems with different practice patterns remains unvalidated.
    The methodological findings (reward-embedded confounding, direct Q
    overestimation) may apply to EHR-based offline RL studies that use
    observational outcomes as rewards.

  \item \textbf{Degenerate MDP.} Our 2-step MDP has 98.3\,\% state invariance
    ($r > 0.99$; of 7 dynamic features, 6 are action one-hot encodings and
    only NIHSS at 24 hours carries clinical information, missing in
    ${\approx}80\,\%$ of records), reducing sequential optimization to
    near-1-step behavior.

  \item \textbf{Over-adjustment risk.} GBM residualization may absorb some
    true treatment-benefit variation. Semi-synthetic calibration suggests
    24--26\,\% absorption, while a conservative propensity-based bound gives
    27\,\%; the adjusted estimate remains clinically small
    ($\vimp_{\text{adj}}\approx+0.0045$, $\leq2.7$\,\% MCID).

  \item \textbf{FQE realizability.} FQE assumes the Q-function class can
    represent the true value function. We cannot rule out realizability
    violations; ensemble-based disagreement tests would strengthen this
    evidence (Appendix~\ref{app:limitations}).

  \item \textbf{Fairness.} T-learner CATE shows non-monotone
    sex$\times$age patterns (Appendix~\ref{app:fairness}), likely from
    extrapolation; these should not be interpreted as demographic
    disparities without prospective validation.
\end{enumerate}

\noindent
Four additional limitations (power at the margin,
outcome endpoint sensitivity, selection bias, post-2018 cohort restriction)
are detailed in Appendix~\ref{app:limitations}.
Action masking enforces clinical guideline constraints (6\,\% violation rate,
all caught at inference; see Appendix~\ref{app:safety}).


\subsection{Recommendations}
\label{sec:discussion:recommendations}

\begin{sloppypar}
For researchers applying offline RL to clinical datasets, we recommend the
6-step empirical evaluation checklist (\S\ref{sec:protocol}) as a set of
pre-specifiable diagnostic checks.
For clinical trialists, ELAN and OPTIMAS narrowed anticoagulation timing, but
largest-infarct estimates remain imprecise and anticoagulation--statin
combinations untested
\citep{fischer2023elan,werring2024optimas,nash2026optimas}.
The T-learner's $4.6\times$ severe/mild CATE ratio therefore motivates stratified
prospective evaluation of this recurrence--hemorrhage trade-off, not treatment
benefit or site-level targeting.
For data infrastructure, clinically useful multi-step RL requires continuous
inpatient monitoring data rather than admission-only registry snapshots.
\end{sloppypar}


\section{Conclusion}
\label{sec:conclusion}

Standard offline RL evaluation for stroke antithrombotic optimization yields
$\vimp=+0.0101$ ($p=0.0002$, E-value $=18.37$), but full deconfounding
reduces it to $+0.0025$ ($p=0.291$), with terminal residualization accounting
for \textbf{218.6\,\%} of the net \rend{}-to-\rfull{} change.
We term this reward-channel pathway \emph{reward-embedded confounding}; external-
confounding E-values do not evaluate it. No family produced positive evaluated evidence
in the common-reward screen; in primary CQL, residualization attenuated the
estimate to null, with T-learner and recurrence AIPW providing triangulation.
Whenever observational outcomes define terminal rewards, positive offline RL
results should remain hypothesis-generating until reward-channel diagnostics
are applied.

%% file: tab_cohort.tex

\begin{table}[h]
  \caption{Baseline characteristics of the post-2018 CRCS-K cohort
           ($N = 44{,}894$).
           Continuous variables are reported as mean $\pm$ SD;
           categorical variables as $n$ (\%).
           \textsuperscript{a}~3-month mRS coverage is 100\,\% in this cohort
           because patients without outcome are excluded from the Phase~8
           MDP dataset (see \S\ref{sec:methods:data}).
           \textsuperscript{b}~TOAST (Trial of Org 10172 in Acute Stroke
           Treatment) classification is missing for 10.5\,\%
           ($n = 4{,}702$); percentages are computed among patients with
           available TOAST ($n = 40{,}192$).
           \textsuperscript{c}~Discharge antithrombotic regimen, classified
           using Config~A (\S\ref{sec:methods:mdp}); AC = anticoagulant,
           AP = antiplatelet.  One patient contributed two admissions;
           the duplicate is assigned to AP\_dual.}
  \label{tab:cohort}
  \centering
  \small
  \begin{tabular}{lc}
    \toprule
    Characteristic & Value \\
    \midrule
    \multicolumn{2}{l}{\textit{Demographics}} \\[2pt]
    Age, years & $68.2 \pm 13.4$ \\
    Female, $n$ (\%) & $17{,}730$ (39.5\,\%) \\[6pt]
    \multicolumn{2}{l}{\textit{Admission severity}} \\[2pt]
    Initial NIHSS score & $4.0 \pm 5.2$ \\[6pt]
    \multicolumn{2}{l}{\textit{Vascular risk factors}} \\[2pt]
    Hypertension, $n$ (\%) & $29{,}602$ (65.9\,\%) \\
    Diabetes mellitus, $n$ (\%) & $14{,}995$ (33.4\,\%) \\
    Atrial fibrillation, $n$ (\%) & $7{,}610$ (17.0\,\%) \\
    Previous stroke, $n$ (\%) & $9{,}332$ (20.8\,\%) \\[6pt]
    \multicolumn{2}{l}{\textit{TOAST stroke aetiology}\textsuperscript{b}} \\[2pt]
    \quad Large-artery atherosclerosis (LAA) & $14{,}951$ (37.2\,\%) \\
    \quad Cardioembolism (CE) & $7{,}383$ (18.4\,\%) \\
    \quad Small-vessel occlusion (SVO) & $7{,}114$ (17.7\,\%) \\
    \quad Incomplete evaluation & $4{,}137$ (10.3\,\%) \\
    \quad Mixed aetiology & $3{,}018$ (7.5\,\%) \\
    \quad Other determined & $1{,}914$ (4.8\,\%) \\
    \quad Undetermined & $1{,}675$ (4.2\,\%) \\[6pt]
    \multicolumn{2}{l}{\textit{Discharge antithrombotic regimen}\textsuperscript{c}} \\[2pt]
    \quad Dual antiplatelet (AP\_dual) & $26{,}430$ (58.9\,\%) \\
    \quad Antiplatelet + statin (AP\_statin) & $6{,}758$ (15.1\,\%) \\
    \quad Anticoagulant + statin (AC\_statin) & $7{,}292$ (16.2\,\%) \\
    \quad Conservative (none) & $2{,}531$ (5.6\,\%) \\
    \quad Anticoagulant alone (AC) & $1{,}122$ (2.5\,\%) \\
    \quad Antiplatelet monotherapy (AP\_mono) & $761$ (1.7\,\%) \\[6pt]
    \multicolumn{2}{l}{\textit{Outcomes}} \\[2pt]
    3-month mRS, mean $\pm$ SD\textsuperscript{a} & $1.7 \pm 1.7$ \\
    Favourable outcome (mRS $0$--$2$), $n$ (\%) & $31{,}739$ (70.7\,\%) \\
    In-hospital death, $n$ (\%) & $744$ (1.7\,\%) \\
    \bottomrule
  \end{tabular}
\end{table}